\documentclass[unnumsec,webpdf,contemporary,large]{oup-authoring-template}%

\graphicspath{{Fig/}}

\usepackage{amsmath}
\usepackage{amsfonts} 
\usepackage[justification=centering]{caption}
\usepackage{booktabs} 
\usepackage{makecell} 

\usepackage{colortbl} 
\usepackage{lineno} 
\usepackage{framed}
\usepackage{color}

\theoremstyle{thmstyleone}

\theoremstyle{thmstyletwo}%
\theoremstyle{thmstylethree}%

\begin{document}

\journaltitle{Journal Title Here}
\DOI{DOI HERE}
\copyrightyear{2023}
\pubyear{2019}
\access{Advance Access Publication Date: Day Month Year}
\appnotes{Paper}

\firstpage{1}

\title[BatmanNet]{BatmanNet: Bi-branch Masked Graph Transformer Autoencoder for Molecular Representation}

\author[1,2]{Zhen Wang}
\author[3]{Zheng Feng}
\author[4]{Yanjun Li}
\author[2]{Bowen Li}
\author[2]{Yongrui Wang}
\author[2]{Chulin Sha}
\author[1,2$\ast$]{Min He}
\author[2,5$\ast$]{Xiaolin Li}

\authormark{Wang et al.}

\address[1]{\orgdiv{College of Electrical and Information Engineering}, \orgname{Hunan University}, \orgaddress{Changsha, \postcode{410082}, \state{Hunan}, \country{China}}}

\address[2]{\orgdiv{Hangzhou Institute of Medicine}, \orgname{Chinese Academy of Sciences}, \orgaddress{Hangzhou, \postcode{310018}, \state{Zhejiang}, \country{China}}}

\address[3]{\orgdiv{Department of Health Outcomes \& Biomedical Informatics, College of Medecine}, \orgname{University of Florida}, \orgaddress{Gainesville, \postcode{32611}, \state{FL}, \country{USA}}}

\address[4]{\orgdiv{Department of Medicinal Chemistry, Center for Natural Products, Drug Discovery and Development}, \orgname{University of Florida}, \orgaddress{Gainesville, \postcode{32610}, \state{FL}, \country{USA}}}

\address[5]{\orgname{ElasticMind Inc}, \orgaddress{Hangzhou, \postcode{310018}, \state{Zhejiang}, \country{China}}}

\corresp[$\ast$]{Corresponding author. \href{email:email-id.com}{hemin607@163.com, xiaolinli@ieee.org}}

\abstract{
Although substantial efforts have been made using graph neural networks (GNNs) for AI-driven drug discovery (AIDD), effective molecular representation learning remains an open challenge, especially in the case of insufficient labeled molecules. Recent studies suggest that big GNN models pre-trained by self-supervised learning on unlabeled datasets enable better transfer performance in downstream molecular property prediction tasks. However, the approaches in these studies require multiple complex self-supervised tasks and large-scale datasets, which are time-consuming, computationally expensive, and difficult to pre-train end-to-end. Here, we design a simple yet effective self-supervised strategy to simultaneously learn local and global information about molecules, and further propose a novel bi-branch masked graph transformer autoencoder (BatmanNet) to learn molecular representations. BatmanNet features two tailored complementary and asymmetric graph autoencoders to reconstruct the missing nodes and edges, respectively, from a masked molecular graph. With this design, BatmanNet can effectively capture the underlying structure and semantic information of molecules, thus improving the performance of molecular representation. BatmanNet achieves state-of-the-art results for multiple drug discovery tasks, including molecular properties prediction, drug-drug interaction, and drug-target interaction, on 13 benchmark datasets, demonstrating its great potential and superiority in molecular representation learning.}

\maketitle

\section{Introduction}
AI-driven drug discovery (AIDD) has attracted increasing research attention. Many remarkable developments have been achieved for the various tasks related to small molecules, e.g., molecular property prediction \citep{ghasemi2018neural}, drug-drug interaction (DDI) prediction \citep{ryu2018deep}, and drug-target (DTI) interaction prediction~\citep{abbasi2021deep,d2020machine,li2019deepatom,li2021dyscore,ye2021unified}, molecule design~\citep{jin2018junction,jin2020hierarchical,wang2021multi}. Effective molecular representation learning plays a crucial role in these downstream tasks. Recently, graph neural networks (GNNs) have exhibited promising potential in this emerging representation learning area, where the atoms and bonds of a molecule are treated as the nodes and edges of a graph \citep{li2020learn}. However, some limitations persist, especially when learning from insufficient labeled molecules, hindering applications to real-world scenarios. In the field of biochemistry, there is a scarcity of task-specific labeled data related to small molecules, primarily due to the high cost and time involved in acquiring high-quality molecular property labels via wet-lab experiments~\citep{xia2022survey}. Supervised training of deep GNNs on these restricted datasets easily leads to the overfitting problem \citep{li2021effective}.

To overcome these challenges, some recent studies suggested that pre-training a large neural network on unlabeled datasets using self-supervised learning enables better transfer performance in downstream molecular property prediction tasks. For example, Sheng et al. \cite{wang2019smiles} and Seyone et al. \cite{chithrananda2020chemberta} used SMILES representation~\citep{weininger1989smiles} of molecules to pre-train a sequence-based model with the masked language-modeling task. However, lacking explicit topology representation, such methods cannot explicitly learn the molecular structural information, instead focusing their learning on the grammar of molecular strings. 
Recently, more research works have proposed to employ self-supervised learning strategies to pre-train models directly from molecular graphs by leveraging large-scale unlabeled molecules \cite{hu2019strategies, rong2020self, li2021effective,li2022kpgt,fang2022geometry}. Although these works achieved better performances on multiple downstream molecular property prediction tasks, we contend that molecular representation learning in this way is suboptimal. In this paper, we argue that current self-supervised learning methodologies applied to molecular graphs continue to confront two principal challenges:

\textbf{Complex pre-training tasks.} Some previous studies have to construct a variety of complex pre-training tasks to learn local and global information about molecules \cite{rong2020self, li2021effective,li2022kpgt,fang2022geometry}. These tasks often require the introduction of additional domain knowledge, such as motifs \cite{rong2020self}, subgraphs \cite{li2021effective}, the atomic distance matrix \cite{fang2022geometry}, molecular descriptors and fingerprints \cite{li2022kpgt}, to manually define the target features predicted during pre-training. The complexity of these pre-training tasks, along with the introduction of additional knowledge, has rendered the entire pre-training process incapable of being effectively accomplished in an end-to-end manner.

\textbf{High computational complexity and large model size.} Given that transformer-based models have demonstrated excellent prediction performance in the fields of NLP and CV, recent molecular pre-training models typically have predominantly adopted transformer-based architectures \cite{rong2020self, li2021effective,li2022kpgt}. Nowadays, transformer-style models typically feature a large number of parameters. Previous molecular pre-training models encoded information directly over the entire molecular graph, resulting in exceedingly high computational complexity and large model size. That is very time-consuming, computationally expensive, and environmentally unfriendly. In addition, large models demand large training datasets and computing resources that might not be readily available, particularly for small research groups or small businesses. Therefore, it is very meaningful to design a simple and effective molecular self-supervised learning strategy while reducing the computational complexity and the number of parameters of the pre-training model.

\begin{figure*}[htbp]
\centering
\includegraphics[width=1.0\textwidth]{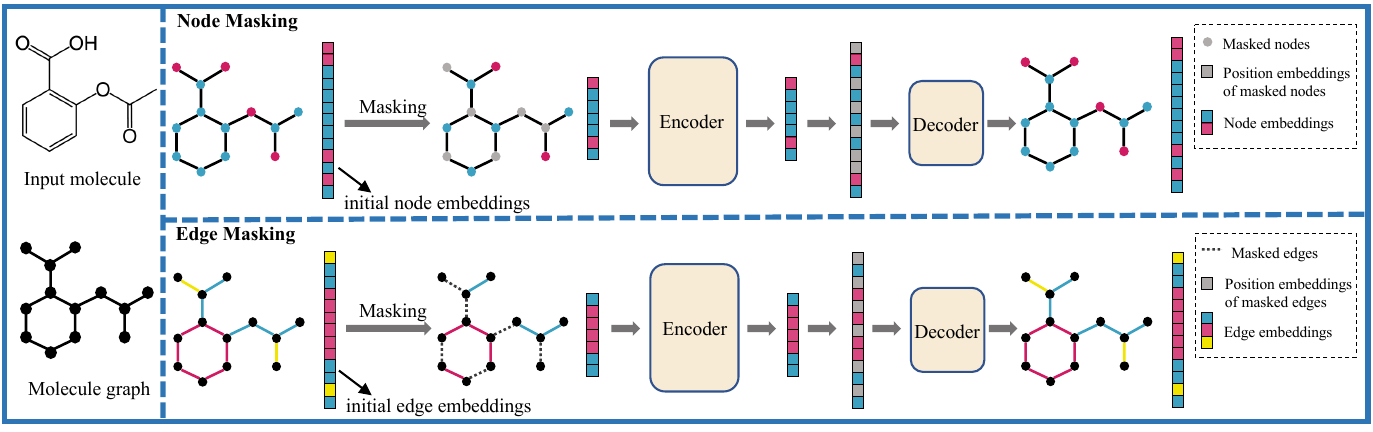} 
\caption{\textbf{Illustration of the designed self-supervised task of BatmanNet.} A very high portion of nodes or edges is randomly masked, and then the BatmanNet is pre-trained to reconstruct the original molecule from the latent representation and mask tokens.}
\label{fig1}
\end{figure*}
 
To address these challenges, we propose a novel molecular self-supervised framework, to alleviate the aforementioned issues and significantly improve the effectiveness and efficiency of molecular representation learning.
\textbf{First}, we introduce a simple yet powerful self-supervised pre-training strategy. Instead of constructing complex pre-training tasks at multiple levels and introducing additional domain-specific chemical knowledge, our strategy is straightforward. We mask a high proportion ($60\%$) of nodes and edges in the molecular graph, respectively, and reconstruct the missing parts through a graph-based autoencoder framework, as illustrated in Figure~\ref{fig1}. This challenging self-supervised task enables our pre-training model to effectively and automatically learn both local and global information about molecular graphs, encouraging the acquisition of expressive structural and semantic knowledge of molecules in an end-to-end fashion. Compared to previous works~\cite {rong2020self, li2021effective, li2022kpgt, fang2022geometry}, our method is significantly more scalable and effective as it directly operates on the finest granularity of atoms and bonds. 
\textbf{Second}, we propose a simple, effective, and scalable form of bi-branch masked graph transformer autoencoder (\textbf{BatmanNet}) for molecular representation learning. Specifically, the encoder is a transformer-style architecture composed of multiple GNN-Attention blocks. The GNN is integrated into the attention layer to extract local and global information of molecular graphs, respectively. BatmanNet has an \textit{asymmetric} encoder-decoder design.
The encoder operates only on the visible subset of the molecular graph (without masked parts). The decoder reconstructs the molecular graph from the learned representation together with masked tokens, and its architecture is similar to the encoder but much more lightweight. 
With this asymmetrical design, the full set of molecular graphs is only processed by the lightweight decoder, which significantly reduces the amount of computation, the overall pre-training time, and memory consumption.

To verify the effectiveness of the proposed BatmanNet, we compared it with several state-of-the-art (SOTA) baselines on a wide range of downstream drug discovery tasks, including molecular properties prediction, drug-drug interaction prediction, and drug-target interaction prediction, with $13$ widely used benchmarks. The experimental results show that our BatmanNet performs much better on multiple drug discovery-related tasks, demonstrating the power capacity, effectiveness, and generalizability of BatmanNet.

In summary, our contributions are as follows:
\begin{itemize}
\item We propose a novel self-supervised pre-training strategy for molecular representation learning to learn both local and global information of the molecules, masking nodes and edges simultaneously with a high mask ratio ($60\%$) and reconstructing them via an autoencoder architecture.
\item We develop a bi-branch asymmetric graph-based autoencoder architecture, significantly enhancing the learning effectiveness and efficiency of the model and vastly reducing memory consumption. 
\item We evaluated BatmanNet thoroughly on various drug discovery tasks. Experimental results demonstrate that BatmanNet outperforms competitive baselines on multiple benchmarks of drug discovery tasks.
\end{itemize}

\section{Related Work}\label{sec2}
\subsection{Molecular Representation Learning}
Many efforts have been devoted to enhancing molecular representation learning to improve the performance of various downstream tasks involving molecules. Early feature-based approaches utilized fixed molecular representations, such as molecular descriptors and fingerprints, to represent molecules in vector spaces \cite{butler2018machine,dong2018admetlab,van2003admet}. However, such methods heavily relied on complex feature engineering to achieve good predictive performance and suffered from vector sparsity issues.

In contrast to feature engineering-based representations, molecular representations learned through deep learning exhibit better generalization and higher expressiveness. Some studies \cite{coley2017convolutional,duvenaud2015convolutional} introduced convolutional layers to learn the neural fingerprints of molecules and applied these neural fingerprints to downstream tasks like property prediction. 
Following these works, \cite{xu2017seq2seq} employed SMILES representations as input and utilized RNN-based models to generate molecular representations. Some works use masked language modeling to pretrain BERT-style models \citep{honda2019smiles,pesciullesi2020transfer,wang2019smiles,chithrananda2020chemberta} or use an autoencoder framework to reconstruct SMILES representations \citep{winter2019learning,gomez2018automatic,xu2017seq2seq}. However, the SMILES itself has several limitations in representing small molecules. First, it is not designed to capture molecular similarity, e.g., two molecules with similar chemical structures might be translated into markedly different SMILES strings, prone to misleading language models with the positional embedding \citep{jin2018junction}. Second, some essential chemical properties of molecules, such as molecular validity, are not readily expressed by the SMILES representation, resulting in more text sequences of invalid molecules. Recently, Graph Neural Networks (GNNs) have been widely applied in learning molecular graph representations, leveraging their significant advantages in modeling graph-structured data. For instance, some works \cite{kearnes2016molecular, schutt2017schnet, schutt2017quantum} explored encoding molecular graphs into neural fingerprints using graph convolutional networks. Work by \cite{ryu2018deeply,xiong2019pushing} proposed learning aggregation weights by extending the Graph Attention Network (GAT) \cite{velivckovic2017graph}. To better capture interatomic interactions, \cite{gilmer2017neural} introduced a message-passing framework, and \cite{gasteiger2020directional,yang2019analyzing} extended this framework to model bond interactions. Additionally, \cite{lu2019molecular} constructed a hierarchical GNN to capture multi-level interactions. While GNNs have made significant progress in the field of molecular graph representation learning, their use of message-passing operators aggregates only local information, making them incapable of capturing long-range dependencies within molecules.

\subsection{Self-supervised Learning for Molecular Graphs}
Self-supervised learning has a long history in machine learning and has yielded fruitful results in many fields, such as computer vision \citep{he2022masked} and language modeling \cite{devlin2018bert}. In light of this influence, self-supervised learning on molecular graphs has emerged as a core direction recently. Current self-supervised learning methods on molecular graphs can be further divided into two subgroups depending on the utilized molecular information level. One group of methods pre-trains $2$D models from the molecular $2$D topology \cite{li2021effective,hu2019strategies,rong2020self,li2022kpgt,liu2019n,chen2022graph,tan2022mgae}. \cite{li2021effective,rong2020self,li2022kpgt} all employ Transformer-style architectures to pre-train molecular graphs. In the case of \cite{li2021effective} and \cite{rong2020self}, motifs or subgraphs from molecules need to be predefined and extracted as prediction targets for their self-supervised tasks. \cite{li2022kpgt}, on the other hand, requires the introduction of additional knowledge, with random masking of a certain proportion of the additional knowledge as part of the reconstruction target. In contrast, our proposed self-supervised learning strategy involves a simple bi-branch graph-masking task that doesn't require specific domain knowledge, such as predefined motifs, subgraphs, or additional information. It is more straightforward, intuitive, and easier to implement. Furthermore, our approach involves the random masking of a high proportion of nodes and edges, each node and edge embedding must learn local contextual information, and the model also needs to learn global information to predict the entire graph from the remaining subgraphs. This makes our task more challenging compared to other self-supervised pretraining tasks. The pre-training models in works \cite{li2021effective,rong2020self,li2022kpgt} are all constructed based on the Transformer architecture, resulting in a large number of model parameters and high computational complexity, necessitating significant computational resources. In contrast, our BatmanNet utilizes an asymmetric transformer-style autoencoder, substantially reducing the model's parameter count and computational complexity, thereby further enhancing the efficiency of our approach.
The other methods worked on the $3$D geometry graphs with spatial positions of atoms by utilizing geometric GNN models~\citep{liu2021pre, fang2022geometry}. Although graph-based methods explicitly consider molecular structural information, they usually require a large volume of molecule data for pre-training due to their complicated architectures, which may limit their generalization abilities when the data is sparse.

Among the graph-based methods, GMAE~\cite{chen2022graph}, MGAE~\cite{tan2022mgae}, and GraphMAE \cite{hou2022graphmae} are the most relevant to our work. However, unlike GMAE masking nodes only and MGAE masking edges only, BatmanNet constructs a bi-branch complementary autoencoder. The dual branches perform node masking and edge masking, respectively, to enhance the expressiveness of the model. GraphMAE is designed to replace masked nodes with descriptors. In contrast, our method directly removes the masked part and adopts two branches to mask nodes and edges, respectively. The learning task in our approach is more challenging than that in GraphMAE, resulting in a more capable model. 
Therefore, we believe that our architecture and self-supervised learning strategy are a superior choice for molecular representation learning compared to other methods.

\section{Materials and methods }\label{sec3}
\subsection{Preliminaries}
\noindent\textbf{Graph Neural Networks (GNNs).}
GNNs are a class of neural networks designed for graph-structured data, and they have been successfully applied in a broad range of domains. 
One of the key components of most GNNs is the message passing (also called neighborhood aggregation) mechanism between nodes in the graph, where the hidden representation $\mathbf{h}_v$ of node $v$ is iteratively updated by aggregating the states of the neighboring nodes and edges. 
For a GNN with $K$ layers, repeating the message passing by $K$ times, the $v$'s hidden representation will contain the structural information of $K$-hop on the graph topology.
Formally, the $k$-th layer of a GNN can be formulated as,

\begin{equation}
\mathbf{m}_{v}^{(k)}=\operatorname{AGG}^{(k)}\left(\left\{\left(\mathbf{h}_{v}^{(k-1)},\mathbf{h}_{u}^{(k-1)}, \mathbf{e}_{u v}\right)\mid u \in \mathcal{N}_{v}\right\}\right),
\label{GNN: message passing}
\end{equation}
\begin{equation}
\mathbf{h}_{v}^{(k)}=\sigma\left(\mathbf{W}^{(k)}\mathbf{m}_{v}^{(k)}+\mathbf{b}^{(k)}\right),
\label{GNN: hv}
\end{equation}

where $\mathbf{m}_{v}^{(k)}$ is the aggregated message, $\mathbf{h}_{v}^{(k)}$ is the representation of node $v$ at the $k$-th layer, $\mathbf{e}_{uv}$ is the representation of edge $(u,v)$, $\sigma(\cdot)$ is the activation function, and $\mathcal{N}_{v}$ is a set neighbors of $v$. $\operatorname{AGG}^{(k)}(\cdot)$ is the neighborhood aggregation process of the $k$-th layer. For convenience, we initialize $\mathbf{h}_{v}^{(0)} = X_v$. After the final iteration $K$, a READOUT function is applied to get the entire graph's representation $\mathbf{h}_G$,

\begin{equation}
\mathbf{h}_{G}=\operatorname{READOUT}\left(\left\{\mathbf{h}_{v}^{\left(K\right)} \mid v \in \mathcal{V} \right\}\right), 
\label{GNN: READOUT}
\end{equation}
where $\mathcal{V}$ is the set of nodes (atoms).

\noindent\textbf{Multi-head attention mechanism.}
The multi-head attention mechanism is the core building block of Transformer~\cite{vaswani2017attention} with several stacked scaled dot-product attention layers. The input of the scaled dot-product attention layer consists of queries $\mathbf{q}$ and keys $\mathbf{k}$ with dimension $d_k$ and values $\mathbf{v}$ of dimension $d_v$. In practice, the set of  ($\mathbf{q}$, $\mathbf{k}$, $\mathbf{v}$)s are packed together into matrices ($\mathbf{Q}$, $\mathbf{K}$, $\mathbf{V}$) so that they can be computed simultaneously. The final output matrix is computed by,
\begin{equation}
    \operatorname{Attention}(\mathbf{Q}, \mathbf{K}, \mathbf{V})=\operatorname{softmax}\left(\frac{\mathbf{Q} \mathbf{K}^{T}}{\sqrt{d_{k}}}\right) \mathbf{V}. 
\label{Attention}
\end{equation}

Multi-head attention allows the model to focus jointly on information from different representation subspaces. Suppose multi-head attention has $h$ parallel attention layers, then the output is,
\begin{equation}
    \operatorname{MultiHead}(\mathbf{Q}, \mathbf{K}, \mathbf{V}) =\operatorname{Concat}\left(\operatorname{head}_{1}, \ldots, \text {head}_{h}\right) \mathbf{W}^{O},
\label{Multi-head attention}
\end{equation}
\begin{equation}
    \text {head}_{i} =\operatorname{Attention}\left(\mathbf{Q} \mathbf{W}_{i}^{\mathbf{Q}}, \mathbf{K} \mathbf{W}_{i}^{\mathbf{K}}, \mathbf{V W}_{i}^{\mathbf{V}}\right),
\label{head}
\end{equation}
where $\mathbf{W}_{i}^{\mathbf{Q}}, \mathbf{W}_{i}^{\mathbf{K}}, \mathbf{W}_{i}^{\mathbf{V}}$ are projection weights of head $i$.

\begin{figure*}[htbp]
\centering
\includegraphics[width=0.78\textwidth]{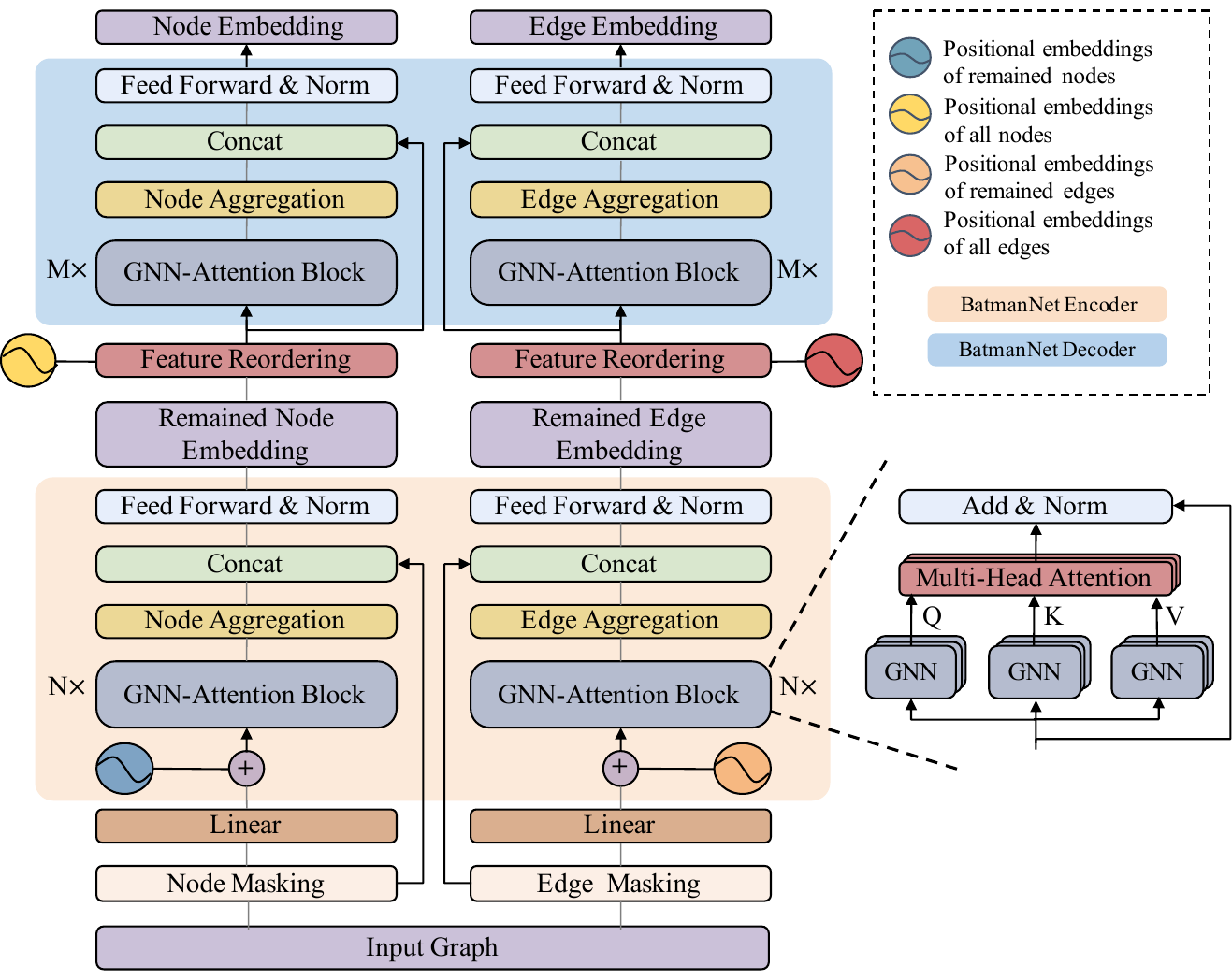} %
\caption{\textbf{Overview of the BatmanNet architecture with node-branch (left) and edge-branch (right)}. The bottom sub-network colored light orange is BatmanNet's encoder. After pre-training, it will be used as a feature extractor for the downstream molecular property prediction tasks by stacking two MLPs on top of two branches, and the final prediction is averaged over the two outputs. The upper sub-network colored light blue is BatmanNet's decoder.}
\vspace{-1em}
\label{fig2}
\end{figure*}

\subsection{Overview of BatmanNet}
This section describes our proposed bi-branch masked graph transformer autoencoder for molecular representation learning (BatmanNet), including the BatmanNet framework and the self-supervised pre-training strategy. 
\subsubsection{The BatmanNet framework}
As depicted in Figure~\ref{fig2}, BatmanNet is a bi-branch model with a node and an edge branch. Each branch focuses on learning the embeddings of nodes or edges from the input graph for fine-tuning downstream tasks. Similar to MAE~\citep{he2022masked}, we propose a transformer-style asymmetric encoder-decoder architecture for each branch. By applying a bi-branch graph masking pre-training strategy, the encoder operates on partially observable signals of molecular graphs and embeds them into latent representations of nodes or edges. The lightweight decoder takes the latent representations of nodes and edges along with mask tokens to reconstruct the original molecule.

For a molecule, we denote the set of nodes (atoms) as $V$ and the set of edges (bonds) as $E$. We introduce node graph $G_N$ and edge graph $G_E$ for each molecule. The node graph is defined as $G_N=(V, E)$, where atom $v \in V$ is regarded as the node of $G_N$ and bond $(u, v) \in E$ as the edge of $G_N$, connecting atoms $u$ and $v$. The initial features of nodes and edges are denoted by $X_v$ and $e_u{}_v$, respectively. We can apply GNNs to the node graph to perform the message passing over nodes. The edge graph $G_E$ is the primary dual of the node graph, describing the neighboring edges in the original graph and ensuring message passing over edges in a similar fashion \citep{chen2017supervised}. The node graph $G_N$ and the edge graph $G_E$ are taken as the inputs of the node branch and edge branch of BatmanNet, respectively.

As shown in Figure \ref{fig2}, the encoder consists of two symmetric multi-layer transformer-styled networks based on the implementation described in \citep{rong2020self}, mapping the initial features of visible, unmasked nodes and edges to embeddings in their latent feature spaces. The decoder takes the input of a complete set of reordered molecular representations, including (i) the embeddings of unmasked nodes and edges from the encoder and (ii) mask tokens of removed nodes and edges. The decoder uses the same transformer-styled architecture as the encoder but is more lightweight, and it is only used in the pre-training stage to perform the molecular reconstruction task. Only the encoder is used to produce molecular representations for the downstream tasks. Based on~\citep{he2022masked}, a narrower or shallower decoder would not impact the overall performance of the MAE. In this asymmetric encoder-decoder design, the nodes and edges of the entire graph are only processed at the lightweight decoder. This significantly reduces the model's computation and memory consumption pre-training. 

\noindent\textbf{Details of encoder and decoder.} 
As illustrated in Figure \ref{fig2}, The encoder and decoder consist of a stack of $N$ and $M$ ($M \ll N$) identical layers of the GNN-Attention block, respectively, with each block adopting a double-layer information extraction framework. The GNN is integrated into the attention layer to extract local and global information of molecular graphs, respectively.

In a GNN-Attention block, a GNN is used as the first layer of the information extraction network. It performs the message passing operation on the input graph to extract local information, and the output is the learned embeddings. A multi-head attention layer is then employed to capture the global information of the graph. Specifically, a GNN-Attention block comprises three GNNs, i.e., $\mathbf{G}_\mathbf{Q}(\cdot), \mathbf{G}_\mathbf{K}(\cdot)$, and $\mathbf{G}_\mathbf{V}(\cdot)$, that learn embedding of queries $\mathbf{Q}$, keys $\mathbf{K}$, and values $\mathbf{V}$ as follows:
\begin{equation}
    \mathbf{Q}= \mathbf{G}_\mathbf{Q}(\mathbf{H}),  \label{GNN-Q}
\end{equation}
\begin{equation}
    \mathbf{K}= \mathbf{G}_\mathbf{K}(\mathbf{H}),  \label{GNN-K}
\end{equation}
\begin{equation}
    \mathbf{V}= \mathbf{G}_\mathbf{V}(\mathbf{H}), \label{GNN-V}
\end{equation}
where $\mathbf{H} \in \mathbb{R}^{n \times d}$ is the hidden representation matrix of $n$ nodes, with an embedding size of $d$. Then we apply equations (\ref{Attention}), (\ref{Multi-head attention}), and (\ref{head}) to obtain the final output of the GNN-Attention block.

At the beginning of the encoder, we use a linear projection with added positional embeddings to preserve the positional information of unmasked nodes and edges. Here we adopt the absolute sinusoidal positional encoding proposed by~\cite{vaswani2017attention}, and the positions of nodes and edges in the input graph are indexed by RDkit before masking.
By feeding the original graph and its dual graph to both branches of the encoder, respectively, we get the aggregated node embedding $\mathbf{m}_v$ and edge embedding $\mathbf{m}_{vw}$ as follows:
\begin{equation}
\mathbf{m}_{v}=\sum_{u \in \mathcal{N}_{v}} \mathbf{h}_{u},
\end{equation}
\begin{equation}
\mathbf{m}_{vw}=\sum_{u \in \mathcal{N}_{v} \backslash w} \mathbf{h}_{uv}.
\end{equation}

We add long-range residual connections from the initial features of nodes and edges to $\mathbf{m}_v$ and $\mathbf{m}_{vw}$ to overcome the vanishing gradient and alleviate over-smoothing at the message passing stage. In the last step, we apply a Feed Forward and LayerNorm to obtain the unmasked node embedding and edge embedding as the final output of the encoder. 

At the beginning of the decoder, we first use a Feature Reordering layer (as shown in Figure~\ref{fig2}) that concatenates the embeddings of unmasked nodes and edges from the encoder and masked tokens of removed nodes and edges, recovering their order in original input graphs by adding corresponding positional embeddings. Afterwards, the decoder uses the same transformer-styled architecture as the encoder to obtain the node embedding and edge embedding. 

\noindent\textbf{Reconstruction Target.} BatmanNet's node and edge branches reconstruct molecules by predicting all features of masked nodes and edges, respectively. The features of node (atom) and edge (bond) we used in BatmanNet are referred to in Supplementary Section 2.1. A linear layer is appended to each decoder's output, and its output dimension is set as the total amount of the feature size of either atoms (for node branch) or bonds (for edge branch). Both the reconstruction tasks of nodes and edges involve high-dimension multi-label predictions, which can alleviate the ambiguity problem discussed by~\citep{rong2020self} where a limited number of the atom or edge types are used as the node/edge level pre-training targets. The pre-training loss is computed on the masked tokens similar to MAE~\citep{he2022masked}, and the final pre-training loss $\mathcal{L}_\text{pre-train}$ is defined as:
\begin{equation}
    \mathcal{L}_\text{pre-train}=\mathcal{L}_\text{node} + \mathcal{L}_\text{edge},
\label{pretrain-loss}
\end{equation}
\begin{equation}
    \mathcal{L}_\text{node}=\sum_{v \in \mathcal{V}_\text{mask}} \mathcal{L}_\text{ce}\left(\boldsymbol{p}_{v}, \boldsymbol{y}_{v}\right),
\label{node-loss}
\end{equation}
\begin{equation}
    \mathcal{L}_\text{edge}=\sum_{(u,v) \in \mathcal{E}_\text{mask}} \mathcal{L}_\text{ce}\left(\boldsymbol{p}_{(u,v)}, \boldsymbol{y}_{(u,v)}\right),
\label{edge-loss}
\end{equation}
where $\mathcal{L}_\text{node}$ and $\mathcal{L}_\text{edge}$ are the loss functions of the node branch and edge branch. $\mathcal{V}_\text{mask}$ and $\mathcal{E}_\text{mask}$ represent the set of masked nodes and edges, respectively.  $\mathcal{L}_\text{ce}$ is the cross entropy loss between predicted node features $\boldsymbol{p}_{v}$ and corresponding ground-truth $\boldsymbol{y}_{v}$ or predicted edge features $\boldsymbol{p}_{(u,v)}$ and corresponding ground-truth $\boldsymbol{y}_{(u,v)}$.

\subsubsection{Pre-training strategy: Bi-branch graph masking}
The efficacy of a pre-trained model heavily relies on the design of self-supervision tasks, which should ideally encompass both node and graph levels to enable the model to learn local and global molecular graph information \citep{li2021effective, hu2019strategies, rong2020self, fang2022geometry}. Inspired by MAE~\citep{he2022masked}, we propose a self-supervised pre-training strategy that accomplishes this goal through a single prediction task, using a bi-branch graph masking and reconstruction approach for molecular pre-training. Specifically, given a molecular graph, our approach randomly masks a high proportion of its nodes and edges in both the node and edge branches of the model. The encoder then operates on the remaining unmasked nodes and edges. It is worth mentioning that, considering that the message passing process in GNNs is directed, we adopt the directed masking scheme~\citep{tan2022mgae} to the random masking of edges (\textit{i.e.}, $(u, v)$ and $(v, u)$ are different). Removing $(u, v)$ does not mean that $(v, u)$ is also removed. To distinguish $(u, v)$ and $(v, u)$, we add the feature of the starting node (head node) to the initial feature of the edge. 

Our design of the strategy is effective for two reasons. First, our node-level pre-training approach enables the learning of local contextual information beyond the k-hops range and limited shapes. Previous node/edge-level pre-training strategies typically rely on multiple prediction tasks to capture domain knowledge by learning neighboring graph structures and the regularities of the node/edge attributes distributed over these graph structures \citep{hu2019strategies} or introducing additional definitions such as motifs \cite{rong2020self}, subgraphs \cite{li2021effective}, and additional knowledge \cite{li2022kpgt}. In contrast, our approach randomly masks a high percentage of nodes and edges, e.g., $60\%$, so that each node/edge has a high likelihood of missing neighboring nodes and edges simultaneously. To reconstruct the missing neighboring nodes and edges, each node and edge embedding must learn its contextual information locally. This high ratio of random masking and reconstruction removes the restriction of scale and shape of subgraphs used for prediction, thus promoting the capturing of local contextual information beyond the k-hops range and limited shapes. Second, our graph-level pre-training approach involves predicting the entire graph from the remaining nodes and edges after random masking, resulting in a more challenging task than other self-supervised pre-training tasks that typically learn global graph information with smaller graphs or motifs as the target \citep{rong2020self, li2021effective}. This more challenging pre-training task of bi-branch graph masking and reconstruction entails a more powerful model with a larger capacity for learning high-quality node and edge embeddings to capture molecular information at both the local and global levels.

Overall, our proposed pre-training strategy achieves a more efficient and effective learning of molecular information at both the node/edge and graph levels while maintaining the learning capacity of the pre-training tasks.

\begin{sidewaystable*}[htbp]
\caption{Overall performance for classification tasks and regression tasks of molecular properties prediction.
}\label{molecular property prediction}
\renewcommand\arraystretch{1.2}
\begin{tabular*}{\textheight}{@{\extracolsep\fill}lccccccccc@{\extracolsep{\fill}}}
\toprule%
\multicolumn{1}{@{}l@{}}{Methods} & \multicolumn{6}{@{}c@{}}{Classification (AUC-ROC)}& \multicolumn{3}{@{}c@{}}{Regression (RMSE)} \\
\midrule
Dataset     & BACE & BBBP & ClinTox & SIDER & Tox21 & ToxCast & FreeSolv & ESOL & Lipo \\
\#Molecules & {1513} & {2039} & {1478}   & {1427}  & {7831} & {8575} & {642} & {1128} & {4200} \\
\#tasks     & {1}  & {1} &  {2}  & {27}    & {12} & {617} &  {1}  & {1}    & {1}  \\
\midrule
ECFP \citep{rogers2010extended}        & 0.861$_{(0.024)}$ & 0.783$_{(0.050)}$ & 0.673$_{(0.031)}$ & 0.630$_{(0.019)}$ & 0.760$_{(0.009)}$ & 0.615$_{(0.017)}$ & 5.275$_{(0.751)}$ & 2.359$_{(0.454)}$ & 1.188$_{(0.061)}$ \\
TF\_Robust \citep{ramsundar2015massively} & 0.824$_{(0.022)}$ & 0.860$_{(0.087)}$ & 0.765$_{(0.085)}$ & 0.607$_{(0.033)}$ & 0.698$_{(0.012)}$ & 0.585$_{(0.031)}$ & 4.122$_{(0.085)}$ & 1.722$_{(0.038)}$ & 0.909$_{(0.060)}$ \\
GraphConv \citep{kipf2016semi}  & 0.854$_{(0.011)}$ & 0.877$_{(0.036)}$ & 0.845$_{(0.051)}$ & 0.593$_{(0.035)}$ & 0.772$_{(0.041)}$ & 0.650$_{(0.025)}$ & 2.900$_{(0.135)}$ & 1.068$_{(0.050)}$ & 0.712$_{(0.049)}$ \\
Weave \citep{kearnes2016molecular}      & 0.791$_{(0.008)}$ & 0.837$_{(0.065)}$ & 0.823$_{(0.023)}$ & 0.543$_{(0.034)}$ & 0.741$_{(0.044)}$ & 0.678$_{(0.024)}$ & 2.398$_{(0.250)}$ & 1.158$_{(0.055)}$ & 0.813$_{(0.042)}$ \\
SchNet \citep{schutt2017schnet}     & 0.750$_{(0.033)}$ & 0.847$_{(0.024)}$ & 0.717$_{(0.042)}$ & 0.545$_{(0.038)}$ & 0.767$_{(0.025)}$ & 0.679$_{(0.021)}$ & 3.215$_{(0.755)}$ & 1.045$_{(0.064)}$ & 0.909$_{(0.098)}$ \\
MPNN \citep{gilmer2017neural}       & 0.815$_{(0.044)}$ & 0.913$_{(0.041)}$ & 0.879$_{(0.054)}$ & 0.595$_{(0.030)}$ & 0.808$_{(0.024)}$ & 0.691$_{(0.013)}$ & 2.185$_{(0.952)}$ & 1.167$_{(0.430)}$ & 0.672$_{(0.051)}$ \\
DMPNN \citep{yang2019analyzing}      & 0.852$_{(0.053)}$ & 0.919$_{(0.030)}$ & 0.897$_{(0.040)}$ & 0.632$_{(0.023)}$ & 0.826$_{(0.023)}$ & 0.718$_{(0.011)}$ & 2.177$_{(0.914)}$ & 0.980$_{(0.258)}$ & 0.653$_{(0.046)}$ \\
MGCN \citep{lu2019molecular}       & 0.734$_{(0.030)}$ & 0.850$_{(0.064)}$ & 0.634$_{(0.042)}$ & 0.552$_{(0.018)}$ & 0.707$_{(0.016)}$ & 0.663$_{(0.009)}$ & 3.349$_{(0.097)}$ & 1.266$_{(0.147)}$ & 1.113$_{(0.041)}$ \\
AttentiveFP \citep{xiong2019pushing} & 0.863$_{(0.015)}$ & 0.908$_{(0.050)}$ & 0.933$_{(0.020)}$ & 0.605$_{(0.060)}$ & 0.807$_{(0.020)}$ & 0.579$_{(0.001)}$ & 2.030$_{(0.420)}$ & 0.853$_{(0.060)}$ & 0.650$_{(0.030)}$ \\
TrimNet \citep{li2021trimnet}    & 0.843$_{(0.025)}$ & 0.892$_{(0.025)}$ & 0.906$_{(0.017)}$ & 0.606$_{(0.006)}$ & 0.812$_{(0.019)}$ & 0.652$_{(0.032)}$ & 2.529$_{(0.111)}$ & 1.282$_{(0.029)}$ & 0.702$_{(0.008)}$ \\
\midrule
\cellcolor{gray!40} Mol2Vec \citep{jaeger2018mol2vec}    & 0.841$_{(0.052)}$ & 0.876$_{(0.030)}$ & 0.828$_{(0.023)}$ & 0.601$_{(0.023)}$ & 0.805$_{(0.015)}$ & 0.690$_{(0.052)}$ & 5.752$_{(1.245)}$ & 2.358$_{(0.452)}$ & 1.178$_{(0.054)}$ \\
\cellcolor{gray!40} N-GRAM \citep{liu2019n}  & 0.876$_{(0.035)}$ & 0.912$_{(0.013)}$ & 0.855$_{(0.037)}$ & 0.632$_{(0.005)}$ & 0.769$_{(0.027)}$ & - & 2.512$_{(0.190)}$ & 1.100$_{(0.160)}$ & 0.876$_{(0.033)}$ \\
\cellcolor{gray!40} SMILES-BERT \citep{wang2019smiles} & 0.849$_{(0.021)}$ & \textbf{0.959}$_{\textbf{(0.009)}}$ & \textbf{0.985}$_{\textbf{(0.014)}}$ & 0.568$_{(0.031)}$ & 0.803$_{(0.010)}$ & 0.665$_{(0.010)}$ & 2.974$_{(0.510)}$ & 0.841$_{(0.096)}$ & 0.666$_{(0.029)}$ \\
\cellcolor{gray!40} pre-trainGNN \citep{hu2019strategies} & 0.851$_{(0.027)}$ & 0.915$_{(0.040)}$ & 0.762$_{(0.058)}$ & 0.614$_{(0.006)}$ & 0.811$_{(0.015)}$ & 0.714$_{(0.019)}$ & - & - & -\\
\cellcolor{gray!40} GraphMAE\footnotemark[1] \citep{hou2022graphmae}   & 0.863$_{(0.002)}$ & 0.896$_{(0.007)}$ & 0.850$_{(0.007)}$ & 0.652$_{(0.001)}$ & 0.794$_{(0.003)}$ & 0.679$_{(0.005)}$ & - & - & - \\
\cellcolor{gray!40} GROVERbase \citep{rong2020self} & 0.878$_{(0.016)}$ & 0.936$_{(0.008)}$ & 0.925$_{(0.013)}$ & 0.656$_{(0.023)}$ & 0.819$_{(0.020)}$ & 0.723$_{(0.010)}$ & 1.592$_{(0.072)}$ & 0.888$_{(0.116)}$ & 0.563$_{(0.030)}$ \\
\cellcolor{gray!40} GROVERlarge \citep{rong2020self} & 0.894$_{(0.028)}$ & 0.940$_{(0.019)}$ & 0.944$_{(0.021)}$ & 0.658$_{(0.023)}$ & 0.831$_{(0.025)}$ & 0.737$_{(0.010)}$ & 1.544$_{(0.397)}$ & 0.831$_{(0.120)}$ & 0.560$_{(0.035)}$ \\
\cellcolor{gray!40} KPGT \citep{li2022kpgt} & 0.855$_{(0.011)}$ & 0.908$_{(0.010)}$ & 0.946$_{(0.022)}$ & 0.649$_{(0.009)}$ & 0.848$_{(0.013)}$ & 0.746$_{(0.002)}$ & 2.121$_{(0.837)}$ & 0.803$_{(0.008)}$ & 0.600$_{(0.010)}$ \\
\cellcolor{gray!40} MPG \citep{li2021effective}       &0.920$_{(0.013)}$ & 0.922$_{(0.012)}$ & 0.963$_{(0.028)}$ & 0.661$_{(0.007)}$ & 0.837$_{(0.019)}$ & 0.748$_{(0.005)}$ & 1.269$_{(0.192)}$ & 0.741$_{(0.017)}$ & \textbf{0.556}$_{\textbf{(0.017)}}$ \\
\cellcolor{gray!40} GEM \citep{fang2022geometry}       & 0.925$_{(0.010)}$ & 0.953$_{(0.007)}$ & 0.977$_{(0.019)}$ & 0.663$_{(0.014)}$ & 0.849$_{(0.003)}$ & 0.742$_{(0.004)}$ & - & - & - \\
\midrule
\cellcolor{gray!40} \textbf{BatmanNet}   & \textbf{0.928}$_{\textbf{(0.008)}}$ & 0.946$_{(0.003)}$ & 0.926$_{(0.002)}$ & \textbf{0.676}$_{\textbf{(0.007)}}$ & \textbf{0.855}$_{\textbf{(0.005)}}$ & \textbf{0.756}$_{\textbf{(0.007)}}$ & \textbf{1.174}$_{\textbf{(0.054)}}$ & \textbf{0.736}$_{\textbf{(0.014)}}$ & 0.578$_{(0.034)}$ \\
\botrule
\end{tabular*}
\begin{tablenotes}
\item The methods in shading cells are pre-trained methods. The SOTA results are shown in bold. Standard deviations are in brackets.
\item[1] To offer a fair comparison, we re-tested GraphMAE\citep{hou2022graphmae} under the same experimental conditions as MPG\citep{li2021effective}.
\end{tablenotes}
\end{sidewaystable*}

\section{Experiments and Results}
To comprehensively evaluate the performance of BatmanNet, we conduct extensive experiments and compare its performance against several SOTA methods across multiple benchmarks. These benchmarks encompass a wide range of molecular property prediction tasks, including those related to physical, chemical, and biophysical properties, as well as drug-drug interaction (DDI) and drug-target interaction (DTI) prediction tasks.

\subsection{Pre-training settings}
\subsubsection{Pre-training Datasets.}
The pre-training of BatmanNet is carried out on the ZINC-$250$K molecule dataset from~\cite{kusner2017grammar}. The dataset is composed of $250$K molecules that were sampled from the ZINC database~\cite{sterling2015zinc}. Here we randomly split the dataset into training and validation sets in a $9:1$ ratio.
\subsubsection{Experimental Configurations.}
We use the Adam~\citep{kingma2014adam} optimizer and the Noam learning rate scheduler~\citep{devlin2018bert} to optimize the model and adjust the learning rate for pre-train. The batch size is set as $32$, and the BatmanNet is implemented by PyTorch \cite{NEURIPS2019_9015}.
The masking ratio for both branches of BatmanNet is set as $0.6$, while the encoder and decoder consist of $6$ and $2$ layers, respectively, with a hidden size of $100$. The GNN-attention block of each layer utilizes $3$ GNN layers and $2$ self-attention heads. The autoencoder structure comprises roughly $2.6M$ parameters and is pre-trained for two days on a single Nvidia RTX3090.
\subsubsection{Pre-trained representations visualization.}
\begin{figure*}[thbp]%
\centering
\includegraphics[width=0.95\textwidth]{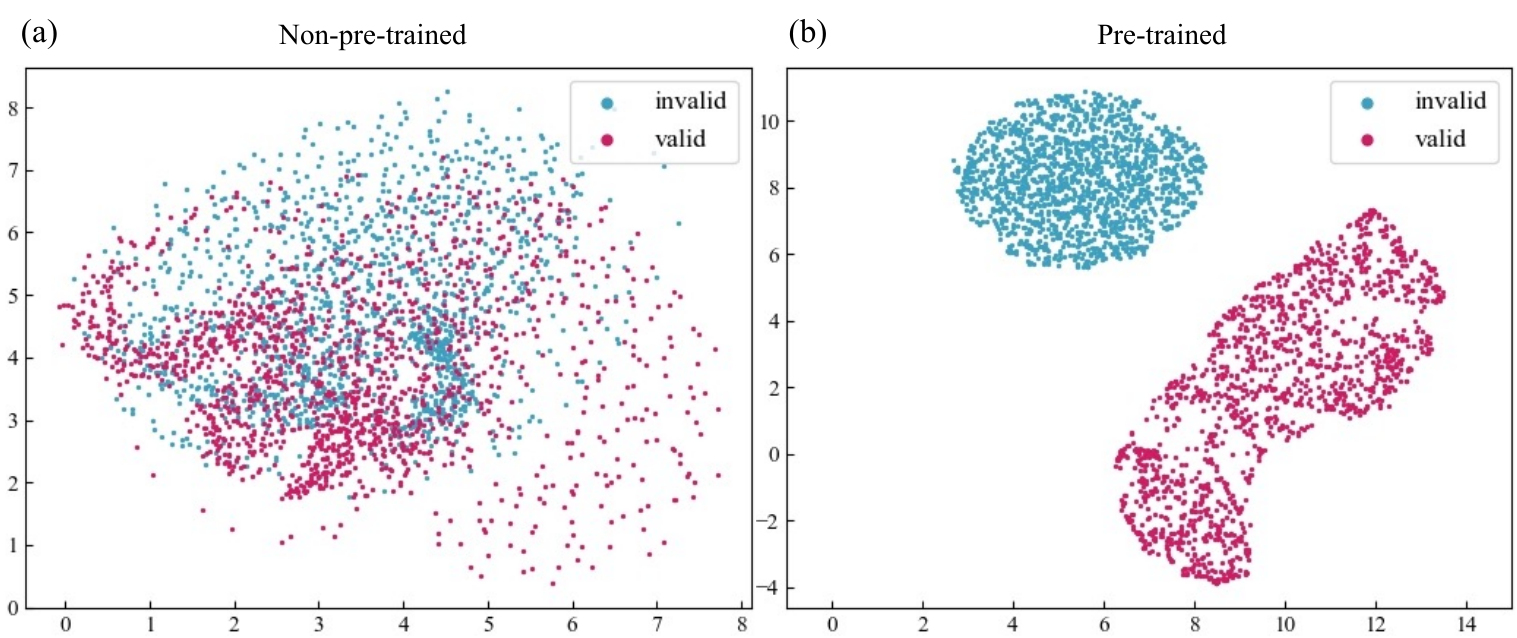}
\caption{\textbf{Visualization of the molecular representation by UMAP.} In \textbf{(a)} and \textbf{(b)}, the molecular representation is the embedding extracted from the last layer of non-pre-trained and pre-trained BatmanNet. The pre-trained BatmanNet is capable of distinguishing molecules with valid structures and those with invalid structures.}
\label{fig: visualization}
\end{figure*}

In order to visually observe the representations that the self-supervised tasks (without downstream fine-tuning) have learned, we projected them into a two-dimensional space for visualization purposes. Here, we investigated whether the pre-training method is able to effectively discriminate between molecules with valid structures and those with invalid structures. We randomly selected 1,500 molecules with valid structures from the ZINC dataset and introduced structural perturbations to generate invalid molecules by shuffling atom features and altering the order of atoms and bonds. For each valid and invalid molecule, we extracted the embedding from the last layer of the pre-trained BatmanNet as molecular representations. Subsequently, we employed the UMAP algorithm \cite{mcinnes2018umap} to map these representations into a two-dimensional space for visualization. We also conducted a similar analysis on the BatmanNet model that was not pre-trained for comparison. As illustrated in Figure \ref{fig: visualization}(a) and  \ref{fig: visualization}(b), in comparison to the not pre-trained model, the pre-trained BatmanNet demonstrates an enhanced ability to distinguish between molecules with valid structures and those with invalid structures, suggesting that pre-trained models can effectively discern the structural validity of molecules.

\subsection{Molecular property prediction settings.}
\subsubsection{Datasets.} 
To assess the efficacy of pre-trained BatmanNet in predicting molecular properties, we conducted experiments on several benchmark datasets from MoleculeNet~\cite{wu2018moleculenet}. These datasets encompass both classification and regression tasks, and additional information about the datasets may be found in Supplementary Section 1.1. Following the previous works\cite{rong2020self, li2021effective, fang2022geometry},  We apply scaffold splitting~\citep{bemis1996properties,ramsundar2019deep} to split the dataset into training, validation, and test sets at a ratio of $8:1:1$ in each downstream task. This approach segregates the dataset into various substructures, enabling the evaluation of the model's ability to generalize outside of the distribution, a challenging yet reliable evaluation. More details are deferred to Supplementary Section 2.
\subsubsection{Baselines.} 
We compare the performance of BatmanNet with $20$ competitive baselines for molecular property prediction. We evaluate $10$ supervised learning models without pre-training, \textit{i.e.}, ECFP \citep{rogers2010extended}, TF\_Robust \citep{ramsundar2015massively}, GraphConv~\citep{kipf2016semi}, Weave~\citep{kearnes2016molecular}, SchNet~\citep{schutt2017schnet}, MPNN \citep{gilmer2017neural}, DMPNN \citep{yang2019analyzing}, MGCN~\citep{lu2019molecular}, AttentiveFP \citep{xiong2019pushing} and TrimNet \citep{li2021trimnet}, and $10$ self-supervised learning models with pre-training, \textit{i.e.}, Mol2Vec \citep{jaeger2018mol2vec}, N-GRAM~\citep{liu2019n}, SMILES-BERT \citep{wang2019smiles}, pre-trainGNN~\citep{hu2019strategies}, GraphMAE \citep{hou2022graphmae}, $\text{GROVER}_\text{base}$, $\text{GROVER}_\text{large}$~\citep{rong2020self}, KPGT \cite{li2022kpgt}, MPG \citep{li2021effective}, and GEM \citep{fang2022geometry}.
Among them, ECFP is a circular topological fingerprint designed for molecular characterization, similarity searching, and structure-activity modeling. TF\_Roubust is a DNN-based multi-task framework that takes molecular fingerprints as input. GraphConv, Weave, and SchNet are three graph convolution models. MPNN and its variants DMPNN and MGCN are models considering edge features during message passing. AttentiveFP is an extension of the graph attention network.
TrimNet is a graph-based approach and employs a novel triplet message mechanism to learn molecular representation efficiently.
Mol2Vec, N-GRAM, and SMILES-BERT are inspired by NLP approaches to pre-train a model on sequential representation. 
pre-trainGNN, GraphMAE, GROVER, and MPG are graph-based pre-training models with various pre-training strategies. 
KPGT is a knowledge-guided pre-training method based on the graph transformer. GEM is a geometry-based graph neural network architecture with several dedicated geometry-level self-supervised learning strategies to learn molecular geometry knowledge. 
We only report classification results for pre-trainGNN, GraphMAE, and GEM since the original implementation does not admit regression tasks without non-trivial modifications.

\subsubsection{Experimental configurations and Evaluation metrics.} 
In order to ensure a fair comparison, we adopted the experimental setup used in previous SOTA methods---GROVER \cite{rong2020self}, MPG \cite{li2021effective}, and GEM \cite{fang2022geometry} (Section 4.1 of GEM's Supplementary Information). We selected the model with the best performance on the validation set. Three independent runs were conducted for each property prediction task, and we reported the mean and standard deviation of the ROC-AUC or RMSE. Further pre-training and fine-tuning details are deferred to Supplementary Section 2.

\subsubsection{Experimental results.} 
Table \ref{molecular property prediction} summarizes the performance of BatmanNet along with previous supervised and self-supervised methods on molecular properties prediction. Our results demonstrate that BatmanNet achieves state-of-the-art performance on 6 out of 9 datasets. Compared to molecular fingerprint and supervised models without pre-training, BatmanNet significantly outperforms them on all datasets. On the classification tasks, BatmanNet exceeds the previous SOTA model---GEM on $4$ out of $6$ datasets. On the regression tasks, BatmanNet outperforms the previous SOTA model---MPG on $2$ out of $3$ datasets. 

\begin{figure}[htbp]
\centering
\includegraphics[width=1.0\columnwidth]{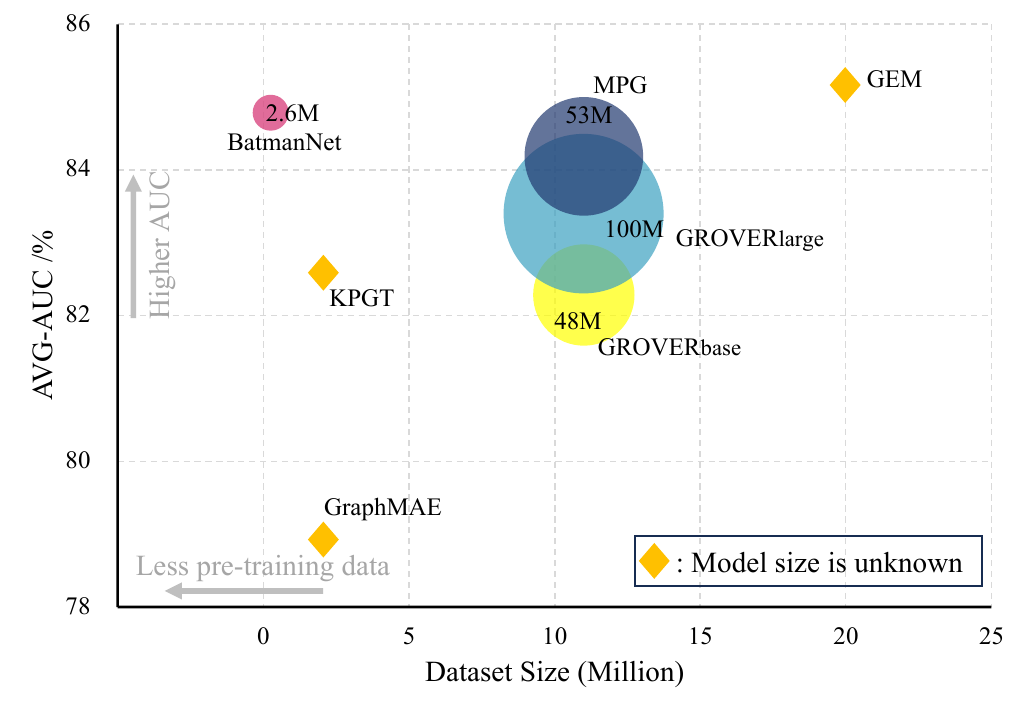} 
\caption{\textbf{Efficacy and Effectiveness Analysis.} The figure illustrates the pre-training dataset size and model size for BatmanNet and a series of advanced baselines, along with their average AUC across all classification datasets about molecular property prediction. The horizontal axis represents the dataset size, the vertical axis represents the average AUC, the circle size corresponds to the model size, and the diamond indicates that the size of the model is unknown.}
\label{fig: model_size}
\end{figure}

\subsubsection{Efficacy and Effectiveness Analysis.}
As illustrated in Figure \ref{fig: model_size}, we conducted a further analysis of the pre-training dataset size and model size for BatmanNet and a series of advanced baselines. We compared their average AUC across all classification datasets about molecular property prediction, see Supplementary Table \ref{tab: model-size}. The results reveal that BatmanNet achieves comparable performance to the previously state-of-the-art models while utilizing fewer training data and model parameters. This finding substantiates the superior efficacy and efficiency of our proposed method.

\begin{table*}[thbp]
\caption{DDI prediction performance of BatmanNet and other various baselines on BIOSNAP dataset  }\label{DDI-biosnap}%
\begin{tabular*}{\textwidth}{@{\extracolsep\fill}lccc}
\toprule
Model & AUC-ROC & PR-AUC & F1 \\
\midrule
LR & 0.802$_{(0.001)}$ & 0.779$_{(0.001)}$ & 0.741$_{(0.002)}$ \\
Nat.Prot \cite{vilar2014similarity} & 0.853$_{(0.001)}$ & 0.848$_{(0.001)}$ & 0.714$_{(0.001)}$ \\
Mol2Vec \cite{jaeger2018mol2vec} & 0.879$_{(0.006)}$ & 0.861$_{(0.005)}$ & 0.798$_{(0.007)}$ \\
MolVAE \cite{gomez2018automatic} & 0.892$_{(0.009)}$ & 0.877$_{(0.009)}$ & 0.788$_{(0.033)}$ \\
DeepDDI \citep{ryu2018deep} & 0.886$_{(0.007)}$ & 0.871$_{(0.007)}$ & 0.817$_{(0.007)}$ \\
CASTER \cite{huang2020caster} & 0.910$_{(0.005)}$ & 0.887$_{(0.008)}$ & 0.843$_{(0.005)}$ \\
GEM \cite{fang2022geometry} & 0.960$_{(0.003)}$ & 0.956$_{(0.002)}$ & 0.903$_{(0.003)}$ \\
MPG \cite{li2021effective} & 0.966$_{(0.004)}$ & 0.960$_{(0.004)}$ & 0.905$_{(0.008)}$ \\
\midrule
BatmanNet & \textbf{0.972}$_{\textbf{(0.001)}}$ & \textbf{0.966}$_{\textbf{(0.001)}}$ & \textbf{0.916}$_{\textbf{(0.002)}}$ \\
\botrule
\end{tabular*}
\begin{tablenotes}%
\item The dataset was divided into training/validation/testing sets in a 7:1:2 ratio. The mean and standard deviation of performances run with three random seeds are reported.
\end{tablenotes}
\end{table*}

\begin{table*}[htbp]
\caption{DTI prediction performance of BatmanNet and other various baselines on Human and C. elegans dataset.}\label{DTI}%
\begin{tabular*}{\textwidth}{@{\extracolsep\fill}llccc}
\toprule
Datasets & Model & Precision & Precision & AUC \\
\midrule
      & Tsubaki et al. \cite{tsubaki2019compound} & 0.923 & 0.918 & 0.970 \\
      & GEM \cite{fang2022geometry} & 0.930 & 0.930 & 0.972 \\
Human & MPG \cite{li2021effective} & 0.952 & 0.940 & 0.985 \\
      & BatmanNet & \textbf{0.983} & \textbf{0.982} & \textbf{0.998} \\
      & (Relative improvement) & (3.26\%) & (4.47\%) & (1.32\%) \\
\midrule
          & Tsubaki et al. \cite{tsubaki2019compound} & 0.938 & 0.929 & 0.978 \\
          & GEM \cite{fang2022geometry} & 0.955 & 0.954 & 0.988 \\
C.elegans & MPG \cite{li2021effective} & 0.954 & 0.959 & 0.986 \\
          & BatmanNet & \textbf{0.988} & \textbf{0.987} & \textbf{0.999} \\
          & (Relative improvement) & (3.46\%) & (2.92\%) & (1.11\%) \\
\botrule
\end{tabular*}
\end{table*}

\begin{table}[thbp]
\caption{5-fold cross-validation DDI prediction performance of BatmanNet and various other baselines on the TWOSIDES dataset, and the results show that BatmanNet outperforms other baselines.}\label{DDI-twosides}%
\begin{tabular*}{\columnwidth}{@{\extracolsep\fill}lccc@{\extracolsep\fill}}
\toprule
Model & Precision & Recall & F1 \\
\midrule
DDI\_PULearn \cite{zheng2019ddi} & 0.904 & 0.824 & 0.862 \\
GEM \cite{fang2022geometry} & 0.928 & 0.929 & 0.928 \\
MPG \cite{li2021effective} & 0.936 & 0.936 & 0.936 \\
\midrule
BatmanNet & \textbf{0.939} & \textbf{0.939} & \textbf{0.939} \\
\botrule
\end{tabular*}
\end{table}

\subsection{Drug-drug interaction prediction settings.}
To assess the effectiveness of BatmanNet on DDI prediction tasks, we formulate the problem as a binary classification task, aiming to predict whether two drugs are likely to interact. Following the approach of MPG \citep{li2021effective}, we compared BatmanNet against recently proposed methods on two real-world datasets---BIOSNAP \citep{zitnik2018biosnap} and TWOSIDES \citep{tatonetti2012data} (For details about both datasets, refer to Supplementary Section 1.1). To ensure a fair comparison, we followed the same experimental procedure as the best approaches---MPG \citep{li2021effective} on the aforementioned datasets. Additionally, to compare with the previous best molecular self-supervised learning model---GEM \citep{fang2022geometry}, we fine-tuned GEM on the two DDI prediction datasets under the same experimental conditions. Classification results are reported in Table \ref{DDI-biosnap} and Table \ref{DDI-twosides}. All baselines, except GEM, are taken from MPG.
Results in Table \ref{DDI-biosnap} and Table \ref{DDI-twosides} indicate that BatmanNet outperforms the previous state-of-the-art method (MPG) on both datasets. These findings demonstrate the superior performance of our BatmanNet on DDI prediction tasks.

\subsection{Drug-target interaction prediction settings.}
Drug-target interaction (DTI) prediction is a critical task in the field of AI-assisted drug discovery (AIDD), aimed at identifying the interaction between a compound and a target protein for drug discovery. Deep learning algorithms have been widely employed for DTI prediction, typically involving encoding compounds and proteins separately. Tsubaki et al. \citep{tsubaki2019compound} proposed a new DTI prediction framework, using GNNs to encode compounds and CNNs to encode proteins. Another recent approach, proposed by Li et al. \citep{li2021effective}, utilized a pre-trained MolGNet as the compound encoder, achieving superior performance compared to existing methods.
In this study, we modified the DTI prediction framework proposed by Tsubaki et al. \citep{tsubaki2019compound}, replacing their compound encoder with our pre-trained BatmanNet to assess its effectiveness. To ensure a fair comparison, we adopted the same experimental procedure as MPG \citep{li2021effective} and evaluated our model on two widely used datasets, namely the Human and C. elegans datasets (Details about both datasets are available in Supplementary Section 1.1). Moreover, to compare the performance of GEM \citep{fang2022geometry} on the DTI prediction task, we employed GEM as the compound encoder and fine-tuned it on the same two datasets.
Results, as shown in Table \ref{DTI}, demonstrate a significant improvement in performance on both datasets when using BatmanNet as the compound encoder, with increases in the precision of 3.26\% and 3.46\%, respectively, compared to the previous best model---MPG \citep{li2021effective}. The results indicate the strong transfer learning capabilities of BatmanNet in molecular representation learning, offering promising prospects for its use in future DTI prediction research.

\begin{figure*}[thbp]%
\centering
\includegraphics[width=1.0\textwidth]{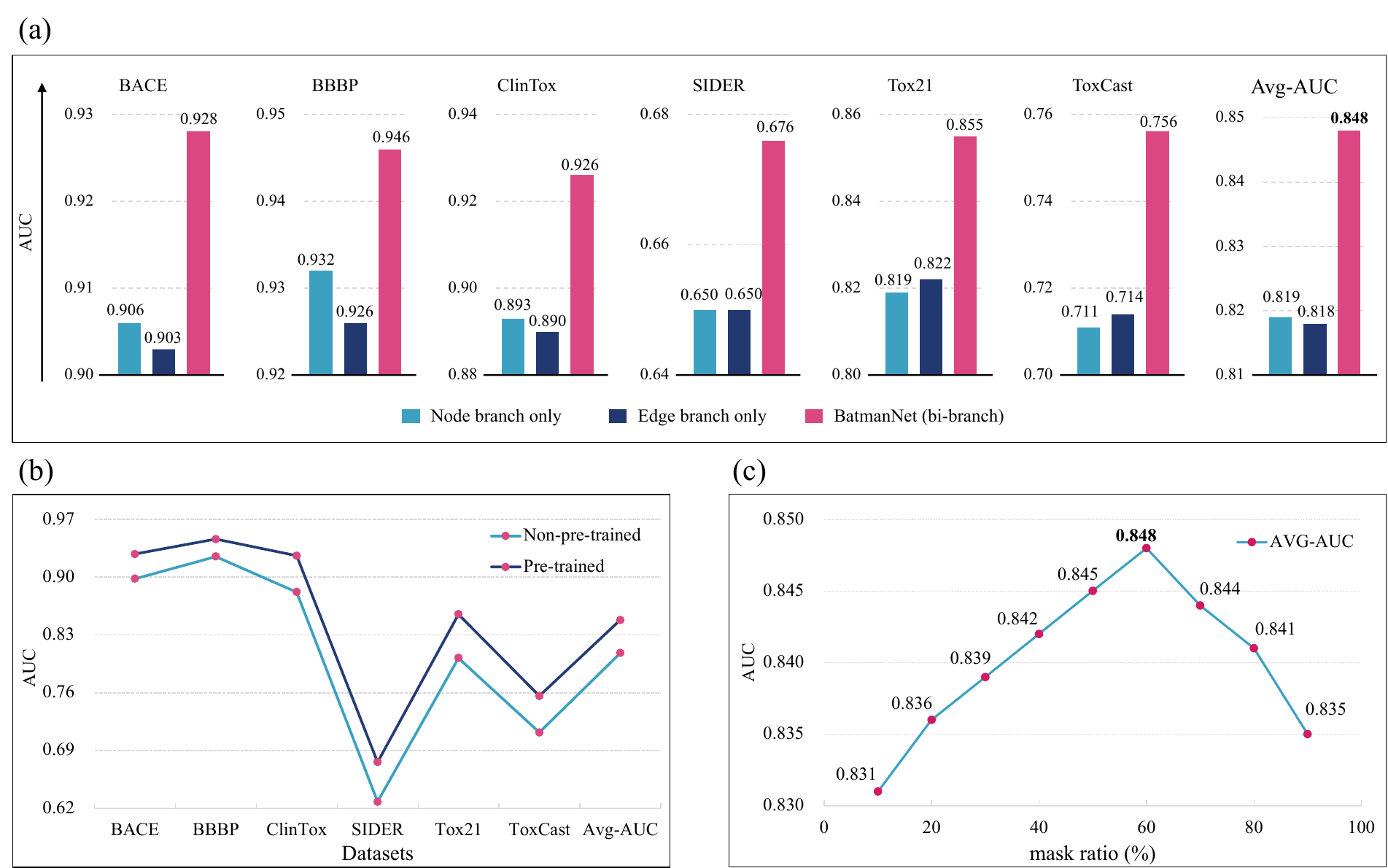}
\caption{\textbf{Ablation studies}. \textbf{(a)}, Comparison between BatmanNet with bi-branch and with a single branch (only node or edge branch) on classification tasks of molecular property prediction. \textbf{(b)}, Comparison results of the pre-trained BaemanNet and the BatmanNet without pre-training on classification tasks of molecular property prediction. \textbf{(c)}, Comparison of different masking ratios. The y-axis is the average AUC on all classification datasets about molecular property prediction in this paper.}
\label{fig: ablation_studies}
\end{figure*}

\subsection{Ablation studies.}
To delve deeper into the factors influencing the performance of our proposed BatmanNet framework, we conducted ablation studies on 6 classification benchmarks about molecular property prediction.

\textbf{Effectiveness of the Bi-branch Information Extraction Network.} To assess the impact of the bi-branch design in BatmanNet, we conducted an ablation study by replacing the bi-branch information extraction network with a single-branch network, either the node branch or the edge branch. Both single-branch networks were pre-trained under the same conditions as the bi-branch network and had nearly the same size of parameters ($2.6M$). Figure~\ref{fig: ablation_studies}(a) shows that using the bi-branch design in BatmanNet improves the average AUC by $2.9\%$ and $3.0\%$ compared to using only the node branch or edge branch, respectively, indicating the effectiveness of the bi-branch design.

\textbf{Impact of the Self-Supervised Pre-training Strategy.}
To evaluate the influence of our self-supervised learning strategy, we compared the classification performance of pre-trained BatmanNet and BatmanNet without pre-training on molecular property prediction tasks using the same hyperparameters. As reported in Figure~\ref{fig: ablation_studies}(b), the pre-trained model consistently outperforms the non-pre-trained model, with an average AUC improvement of $4.0\%$. These findings indicate that our self-supervised pre-training strategy effectively captures rich structural and semantic information about molecules, leading to performance improvement in downstream tasks.

\textbf{Effect of Different Masking Ratio.} To examine the impact of different masking ratios, we conducted a study to pre-train BatmanNet with masking ratios ranging from $10\%$ to $90\%$, then measured the average AUC in all downstream tasks. The detailed results of this study are available in Supplementary Section 3.2, while Figure~\ref{fig: ablation_studies}(c) provides an overview of our findings. We observed that setting the masking ratio to $60\%$ led to the best performance. These results suggest that when a relatively high masking ratio is used, nodes and edges are masked with a greater frequency, resulting in a more challenging pre-training task. This challenging task provides a larger capacity for remaining nodes and edges to capture molecular information in their embeddings. However, when the masking ratio was set higher than $60\%$, there was not enough information in the remaining nodes and edges to recover the complete graph. As a result, the quality of learned embeddings started to decline.

\section{Discussion}
Effective molecular representation learning holds significant importance in the field of AIDD and plays a crucial role in various downstream tasks involving molecules. In this paper, we address two major challenges in current self-supervised learning methods for molecular graphs: complex pre-training tasks and high computational complexity with large model sizes. We introduce a novel self-supervised learning framework for molecular representation learning, consisting of the bi-branch masked graph transformer autoencoder (BatmanNet) and a bi-branch graph masking self-supervised learning strategy. This framework aims to mitigate the mentioned issues, facilitating the learning of more generalizable, transferable, and robust representations of molecular graphs. Extensive test results demonstrate that BatmanNet outperforms current state-of-the-art methods across multiple molecular downstream task datasets.

Our approach offers advantages over several previously advanced methods. \textbf{First}, prior work on molecular self-supervised learning often required constructing multiple tasks to learn local and global information about molecules \cite{hu2019strategies, rong2020self, li2021effective, fang2022geometry}, along with the need of predefined additional domain knowledge, such as motifs \cite{rong2020self}, subgraphs \cite{li2021effective}, the atomic distance matrix \cite{fang2022geometry}, or additional knowledge \cite{li2022kpgt}. Our method only necessitates a simple bi-branch graph masking task combined with a customized pre-training architecture (BatmanNet) to simultaneously learn local and global information about molecules, without requiring additional domain knowledge. This makes our method more scalable and easily applicable to larger pre-training datasets, enhancing the generalization ability of the pre-trained model. It is also more general and can be adapted to graph representation learning in other fields.

\textbf{Second}, most previous molecular pre-trained models relied on the Transformer architecture \cite{rong2020self, li2021effective, fang2022geometry}, directly encoded information from the entire molecular graph, leading to high computational complexity and a large number of model parameters. In contrast, our asymmetric transformer-based encoder-decoder architecture, where the encoder operates on partially observable signals of the molecular graph, processing the complete molecular graph only at the lightweight decoder stage, significantly reduces the model's parameters and computational complexity, further enhancing efficiency. Comparative performance evaluations with extensively trained models reveal that our method achieves superior results with fewer pre-training data and model parameters.

While BatmanNet can generate more effective molecular representations, several potential enhancements are worth exploring. \textbf{First}, consideration of three-dimensional structural information in molecules. The current approach primarily focuses on the planar topological structure of molecules, neglecting the three-dimensional structural information. By incorporating three-dimensional structural details into node and edge features, the model can achieve a more comprehensive understanding of molecular features. This enhancement facilitates the accurate capture of three-dimensional structural information, thereby improving model performance, particularly in tasks involving three-dimensional structures, such as drug-target interactions. \textbf{Second}, expansion of pre-training dataset size. Due to computational constraints, the current small-scale models and datasets may limit the model's performance. Further expanding our approach on a larger pretraining dataset could be valuable in assessing how much improvement can be achieved with our current small-scale models and pretraining dataset. \textbf{Third}, mitigating inherent biases in data-driven approaches. The current completely data-driven approach may exhibit potential data biases. Exploring effective strategies to integrate domain knowledge into our current data-driven pipelines on a primarily data-driven basis to mitigate this bias and improve model performance will be the focus of our future efforts.

In summary, our work underscores the importance of designing effective pre-training tasks for molecular representation learning and demonstrates the effectiveness of a simple, scalable, and domain-agnostic approach based on autoencoding tasks. We believe that our approach has great potential for improving the performance of molecular representation learning and can be applied to a wide range of downstream tasks in the future.

\begin{framed}
\section{Key Points}
\begin{itemize}
\item We develop a novel deep graph neural network--BatmanNet, significantly enhancing the learning effectiveness and efficiency of the model and vastly reducing memory consumption. 
\item A unique self-supervised pre-training strategy is proposed to train BatmanNet, enabling it to learn both local and global information about molecules effectively.
\item BatmanNet consistently outperforms current state-of-the-art methods across diverse drug discovery tasks, including molecular properties prediction, drug-drug interaction, and drug-target interaction.
\end{itemize}
\end{framed}

\section{Data Availability}
The pre-training datasets are available on the ZINC (\url{http://zinc15.docking.org/}). The molecular properties data supporting this study's findings are available on the website of MoleculeNet: \url{http://moleculenet.ai}. The DDI data sets including BIOSNAP and TWOSIDES are available at CASTER repository (\url{https://github.com/kexinhuang12345/CASTER}) and DDI-PULearn additional files (\url{https://drive.google.com/drive/folders/1wKnY4L4iAjBdTMcJBewYNqCgUQ15DXmYusp=sharing}). The DTI data sets including Human and C. elegans are available at \url{https://github.com/masa shitsubaki/CPI\_prediction}.

\section{Code Availability}
The source code is available on GitHub: \url{https://github.com/wz-create/BatmanNet}.

\section{Author Contribution}
Z.W. and X.L. conceived the research project. X.L., M.H., Z.F., and Y.L. supervised and advised the research project. Z.W. designed and implemented the BatmanNet framework. Z.W., Z.F., Y.L., and B.L. conducted the computational analyses. Z.W., Z.F., Y.L., and X.L. wrote the manuscript. All the authors discussed the experimental results and commented on the manuscript.

\section{Acknowledgments}
The authors thank the anonymous reviewers for their valuable suggestions. 

\section{Funding}
This work is supported in part by funds from the National Key Research and Development Program of China (2022YFC3600902) and the Zhejiang Province Soft Science Key Project (2022C25013).

\section{Competing Interests Statement}
The authors declare no competing interests.

\bibliographystyle{unsrt} 
\bibliography{reference}

\clearpage
\begin{appendices}

\section{Supplementary Section 1: Details about Molecular Datastes}
\subsection{1.1 Downstream task datasets.}
In this paper, we have evaluated our method on a wide range of downstream drug discovery tasks, including molecular properties prediction, drug-drug interaction (DDI) prediction, and drug-target interaction (DTI) prediction.

\textbf{Molecular properties prediction datasets}
\begin{itemize}
\item \textbf{BBBP} \citep{bbbp} provides records of whether a compound carries the permeability property of penetrating the blood-brain barrier.
\item \textbf{SIDER} \citep{sider} is a database of marketed drugs and adverse drug reactions (ADR), grouped into 27 system organ classes.
\item \textbf{ClinTox} \citep{clintix} compares drugs approved through the FDA and drugs eliminated due to toxicity during clinical trials.
\item \textbf{BACE} \citep{bace} provides quantitative blinding results for a set of inhibitors of human $\beta$-secretase 1 (BACE-1).
\item \textbf{Tox21} \citep{tox21} is a public database measuring the toxicity of compounds on 12 different targets, including nuclear receptors and stress response.
\item \textbf{ToxCast} \citep{toxcast} providing toxicology data for 8615 compounds based on in vitro high-throughput screening.
\item \textbf{FreeSolv} \citep{freesolv} provides experimental and calculated hydration free energy of small molecules in water. The calculated values are derived from alchemical free energy calculations using molecular dynamics simulations.
\item \textbf{ESOL} \citep{esol} is a small dataset consisting of water solubility data for 1128 compounds. 
\item \textbf{Lipo} \citep{lipo} is curated from the ChEMBL database, which is an important feature of drug molecules that affects both membrane permeability and solubility and provides experimental results of octanol/water distribution coefficient (log D at pH 7.4) of 4200 compounds.
\item \textbf{QM7} \citep{qm7} is a subset of the GDB-13 database, a database of nearly 1 billion stable and synthetically accessible organic molecules, containing up to seven “heavy” atoms (C, N, O, S). 
\item \textbf{QM8} \citep{qm8} are applied to a collection of molecules that include up to eight heavy atoms (also a subset of the GDB-17 database). It contains computer-generated quantum mechanical properties.
\end{itemize}

\textbf{DDI prediction datasets}
\begin{itemize}
\item \textbf{BIOSNAP} \citep{zitnik2018biosnap} that consists of 1322 approved drugs with 41520 labeled DDIs, obtained through drug labels and scientific publications.
\item \textbf{TWOSIDES} \citep{tatonetti2012data} contains side effects caused by the combination of drugs, which contains 548 drugs and 48584 pair-wise drug-drug interactions.
\end{itemize}

\textbf{DTI prediction datasets}
\begin{itemize}
\item \textbf{Human} and \textbf{C.lelgan}, created by Liu et al., include highly credible negative samples of compound-protein pairs by using a systematic framework. Positive samples of the datasets were retrieved from DrugBank 4.1 and Matador. We used a balanced dataset with a ratio of 1:1 of positive and negative samples following Tsubaki et al. \citep{tsubaki2019compound} and MPG \citep{}{li2021effective}.
\end{itemize}

\textbf{Dataset Splitting.} 
In most machine learning applications, the traditional method of random splitting is used to split the dataset. However, in practice, the molecules used for testing may be different from the training molecules in scaffold structure, i.e., out-of-distribution prediction and the way of random splitting is ideal for simulating real-world situations. Unlike random splitting, scaffold splitting splits the data set into different subsets according to the molecule's structure. This challenging but reliable splitting method tests the model's generalization ability outside the distribution (out-of-distribution generalization). We use scaffold splitting to split the dataset into training, validation, and test sets at a ratio of 8:1:1 in each downstream task.

\section{Supplementary Section 2: Implementation Details}
\subsection{2.1 Atom and bond features}
We use RDKit to extract the atom and bond features as the input of GNN and the reconstruction target of BatmanNet. Table \ref{features} shows the atom and bond features we used in BatmanNet.
\subsection{2.2 Pre-training Details}
We use the Adam optimizer with an initial learning rate of 0.0002 and L2 weight decay for $10^{-7}$. We train the model for 20 epochs. The learning rate warmed over the first epoch and decreased exponentially from 0.0004 to 0.0001. Table \ref{tab: pre-train hyper-parameters} demonstrates all the hyper-parameters of the pre-training model, Among these, the parameter mask\_ratio is chosen based on the experimental results in Table \ref{tab: mask ratio}, and the batch\_size is selected according to the GPU memory. The parameters depth, num\_enc\_mt\_block, num\_dec\_mt\_block, num\_dec\_mt\_block, and num\_attn\_head are determined based on the model parameter amount, convergence situation and experience.

\subsection{2.3 Fine-tuning Details}
\subsubsection{Fine-tuning implementation}
We only use BatmanNet's encoder for downstream tasks. Unlike the pre-training, where the model input is an incomplete molecule, the inputs of downstream tasks are complete molecules without masking. After $N$ GNN-Attention blocks, both branches of BatmanNet's encoder perform Node Aggregation, producing two node representations $\mathbf{m}_v^\text{node-branch}$ and $\mathbf{m}_v^\text{edge-branch}$ as follows:
\begin{equation}
\mathbf{m}_v^\text{node-branch}=\sum_{u \in \mathcal{N}_{v}} \overline{\mathbf{h}}_{u}, 
\end{equation}
\begin{equation}
\mathbf{m}_v^\text{edge-branch}=\sum_{u \in \mathcal{N}_{v} \backslash w} \overline{\mathbf{h}}_{uv},  
\end{equation}
where $\overline{\mathbf{h}}_{u}$ and  $\overline{\mathbf{h}}_{uv}$ are the hidden states of the GNN-Attention blocks of node-branch and edge-branch. Then we also apply a single long-range residual connection to concatenate $\mathbf{m}_v^\text{node-branch}$ and $\mathbf{m}_v^\text{edge-branch}$ with initial node features and edge features, respectively. Finally, we transform the two embeddings $\mathbf{m}_v^\text{node-branch}$ and $\mathbf{m}_v^\text{edge-branch}$ through Feed Forward layers and LayerNorm to generate the final two embeddings output for downstream tasks.

Through the above process, given a molecule $G_i$ and the corresponding label $\boldsymbol{y}_i$, BatmanNet's encoder can generate two node embeddings, $\mathbf{H}_i^\text{node-branch}$ and $\mathbf{H}_i^\text{edge-branch}$, from the node branch and the edge branch, respectively. Following GROVER~\citep{rong2020self}, we feed these two node embeddings into a shared self-attentive READOUT function to generate two graph-level embeddings, $\boldsymbol{g}^\text{node-branch}$ and $\boldsymbol{g}^\text{edge-branch}$. They are both obtained by:
\begin{equation}
\mathbf{S} =\operatorname{softmax}\left(\mathbf{W}_{2} \tanh \left(\mathbf{W}_{1} \mathbf{H}^{\top}\right)\right), 
\label{self-attentive READOUT-S}
\end{equation}
\begin{equation}
\boldsymbol{g} =\operatorname{Flatten}(\mathbf{S H}),
\label{self-attentive READOUT-g}
\end{equation}
where $\mathbf{W}_{1} \in \mathbb{R}^{d_\text{attn\_hidden} \times d_\text {hidden\_size}}$ and $\mathbf{W}_{2} \in \mathbb{R}^{d_\text{att\_out} \times d_\text{att\_hidden }}$ are two weight matrix. 

\textbf{Molecular properties prediction.}
After getting two graph-level embeddings $\boldsymbol{g}^\text{node-branch}$ and $\boldsymbol{g}^\text{edge-branch}$, we apply a Feed Forward layer for both branches to get predictions $\boldsymbol{p}_i^\text{node-branch}$ and $\boldsymbol{p}_i^\text{edge-branch}$.
\begin{equation}
\boldsymbol{p}_i^\text{node-branch} = f\left(\boldsymbol{W} \boldsymbol{g}^\text{node-branch}+b\right),
\label{property predictions p-node}
\end{equation}
\begin{equation}
\boldsymbol{p}_i^\text{edge-branch} = f\left(\boldsymbol{W} \boldsymbol{g}^\text{edge-branch}+b\right),
\label{property predictions p-edge}
\end{equation}

\textbf{DDI prediction.} 
In the DDI prediction task, the input is a pair of molecules, which are encoded into two sets of graph-level embeddings ($\boldsymbol{g}_1^\text{node-branch}$ and $\boldsymbol{g}_1^\text{edge-branch}$, $\boldsymbol{g}_2^\text{node-branch}$ and $\boldsymbol{g}_2^\text{edge-branch}$) by the encoder of BatmanNet, respectively. The predictions $\boldsymbol{p}_i^\text{node-branch}$ and $\boldsymbol{p}_i^\text{edge-branch}$ are calculated by:
\begin{equation}
\boldsymbol{p}_i^\text{node-branch} = f\left(\boldsymbol{W} \boldsymbol{g}_{pair}^\text{node-branch}+b\right),
\label{DDI predictions p-node}
\end{equation}
\begin{equation}
\boldsymbol{p}_i^\text{edge-branch} = f\left(\boldsymbol{W} \boldsymbol{g}_{pair}^\text{edge-branch}+b\right),
\label{DDI predictions p-edge}
\end{equation}
\begin{equation}
\boldsymbol{g}_{pair}^\text{node-branch} = \operatorname{Concat}\left(\boldsymbol{g}_1^\text{node-branch}, \boldsymbol{g}_2^\text{node-branch}\right),
\label{DDI predictions g-node}
\end{equation}
\begin{equation}
\boldsymbol{g}_{pair}^\text{edge-branch} = \operatorname{Concat}\left(\boldsymbol{g}_1^\text{edge-branch}, \boldsymbol{g}_2^\text{edge-branch}\right),
\label{DDI predictions g-edge}
\end{equation}

\textbf{DTI prediction.} 
In this study, following MPG \citep{li2021effective}, we adapt Tsubaki et al.'s DTI framework to accomplish the DTI prediction task by replacing the molecular encoder (GNN) with our BatmanNet's encoder. The protein sequence encoder is a CNN model. It uses the attention mechanism to derive the protein sequence representation $\boldsymbol{y}_p$. Given a set of hidden vectors of sub-sequences in a protein $S = (s_1, s_2, ..., s_n)$, the $\boldsymbol{y}_p$ is calculated by:
\begin{equation}
\boldsymbol{y}_{p} =\sum_{i}^{n}\left(\alpha_{i} h_{i}\right),
\label{protein vector yp}
\end{equation}
\begin{equation}
\boldsymbol{\alpha}_{i} =\sigma\left(\boldsymbol{h}_{m}^{T} \boldsymbol{h}_{i}\right), 
\label{protein vector ai}
\end{equation}
\begin{equation}
\boldsymbol{h}_{m} =f\left(\boldsymbol{W} \boldsymbol{g}_{m}+b\right),  
\label{protein vector hm}
\end{equation}
\begin{equation}
\boldsymbol{h}_{i} =f\left(\boldsymbol{W} s_{i}+b\right)
\label{protein vector hi}
\end{equation}
where $\boldsymbol{g}_{m}$ is a molecular vector and the weight for $s_i$ considering $\boldsymbol{g}_{m}$. $\boldsymbol{W}$ is the pearned weight matrix, $b$ is the bias vector, and $\boldsymbol{\alpha}_{i}$ is the attention weights.

Like the molecular properties prediction, we get two molecular embeddings $\boldsymbol{g}^\text{node-branch}$ and $\boldsymbol{g}^\text{edge-branch}$ by the encoder of BatmanNet. Then we get two protein embeddings $\boldsymbol{y}_p^\text{node-branch}$ and $\boldsymbol{y}_p^\text{edge-branch}$ by formula (\ref{protein vector yp}). The predictions $\boldsymbol{p}_i^\text{node-branch}$ and $\boldsymbol{p}_i^\text{edge-branch}$ are calculated by:
\begin{equation}
\boldsymbol{p}_i^\text{node-branch} = f\left(\boldsymbol{W} \boldsymbol{y}_{pair}^\text{node-branch}+b\right),
\label{DTI predictions p-node}
\end{equation}
\begin{equation}
\boldsymbol{p}_i^\text{edge-branch} = f\left(\boldsymbol{W} \boldsymbol{y}_{pair}^\text{edge-branch}+b\right),
\label{DTI predictions p-edge}
\end{equation}
\begin{equation}
\boldsymbol{y}_{pair}^\text{node-branch} = \operatorname{Concat}\left(\boldsymbol{g}^\text{node-branch}, \boldsymbol{y}_p^\text{node-branch}\right),
\label{DTI predictions y-node}
\end{equation}
\begin{equation}
\boldsymbol{y}_{pair}^\text{edge-branch} = \operatorname{Concat}\left(\boldsymbol{g}^\text{edge-branch}, \boldsymbol{y}_p^\text{edge-branch}\right),
\label{DTI predictions y-edge}
\end{equation}

\textbf{The final loss} of downstream tasks consists of the supervised loss $\mathcal{L}_\text{sup}$ and the disagreement loss \citep{li2019semi} $\mathcal{L}_\text{diss}$, where the disagreement loss is to train the two predictions to be consistent. 
\begin{equation}
\mathcal{L}_\text{fine-tune} = \mathcal{L}_\text{sup} + \mathcal{L}_\text{diss},
\label{fine-tune loss}
\end{equation}
\begin{equation}
\mathcal{L}_\text{sup} = \mathcal{L}\left(\boldsymbol{p}_i^\text{node-branch}, \boldsymbol{y}_{i}\right)+\mathcal{L}\left(\boldsymbol{p}_i^\text {edge-branch}, \boldsymbol{y}_{i}\right), 
\label{sup loss}
\end{equation}
\begin{equation}
\mathcal{L}_{\text{diss}} =\left\|\boldsymbol{p}_i^\text {node-branch}-\boldsymbol{p}_i^\text{edge-branch}\right\|_{2}.
\label{diss loss}
\end{equation}

\subsubsection{The Fine-tuning Hyperparams}
For each task, we try different hyper-parameter combinations via random search to find the best results Table \ref{tab: fine-tune hyper-parameters} shows all the hyper-parameters of the fine-tuning model.

\setcounter{table}{0}
\renewcommand{\thetable}{S\arabic{table}}
\begin{table*}[htb]
\centering
\caption{Atom and Bond features.}
\begin{tabular*}{\textwidth}{@{\extracolsep{\fill}}llll@{\extracolsep{\fill}}}
    \toprule
         &Features & Size & Description \\ \midrule
         &Atom type & 23 & The atom type (e.g., C, N, O), by atomic number\\ 
         &Number of H & 6 & The number of bonded hydrogen atoms \\ 
    Atom &Charge & 5 & The formal charge of the atom\\ 
         &Chirality & 4 & The chiral-tag of the atom\\ 
         &Is-aromatic & 1 & Whether the atom is part of an aromatic system or not\\ \midrule
    Bond &Bond type & 5 &  The bond type (e.g., single, double, triple et al.)\\
         &Stereo & 6 & The stereo-configuration of the bond\\ 
     \bottomrule
\end{tabular*}
\label{features}
\end{table*}

\begin{table*}[htb]
\centering
\caption{The pre-training hyper-parameters.}
\begin{tabular*}{\textwidth}{@{\extracolsep\fill}lll}
\toprule
Hyper-parameter & Value & Description \\ \midrule
batch\_size & 32 & The input batch\_size\\ 
hidden\_size & 100 & The hidden\_size of encoder and decoder \\ 
depth & 3 & The number of GNN layers in GNN-Attention block\\ 
num\_enc\_mt\_block & 6 & The number of the GNN-Attention block in encoder\\ 
num\_dec\_mt\_block & 2 & The number of the GNN-Attention block in decoder\\ 
num\_attn\_head & 2 & The number of attention heads in the GNN-Attention block\\
mask\_ratio & 0.6 &  The mask ratio\\
init\_lr & 0.0002 & The initial learning rate of Noam learning rate schedular\\ 
max\_lr & 0.0004 & The maximum learning rate of Noam learning rate schedular\\ 
final\_lr & 0.0001 & The final learning rate of Noam learning rate schedular\\ 
\bottomrule
\end{tabular*}
\label{tab: pre-train hyper-parameters}
\end{table*}

\begin{table*}[htb]
\centering
\caption{The fine-tuning hyper-parameters.}
\begin{tabular*}{\textwidth}{@{\extracolsep\fill}lll}
\toprule
Hyper-parameter & Value & Description \\ \midrule
batch\_size & 32 & The input batch\_size\\ 
ffn\_hidden\_size & 200 & The hidden\_size of MLP layers \\ 
ffn\_num\_layer & 2 & The number of MLP layers\\ 
attn\_hidden & 200 & The hidden\_size for the self-attentive readout\\ 
attn\_out & 2 & The number of output heads for the self-attentive readout \\ 
dist\_coff & 0.1 &  The coefficient of the disagreement loss\\
init\_lr & max\_lr / 10 & The initial learning rate of Noam learning rate schedular\\ 
max\_lr & 0.0001~0.001 & The maximum learning rate of Noam learning rate schedular\\ 
final\_lr & max\_lr / (5-10) & The final learning rate of Noam learning rate schedular\\ 
\bottomrule
\end{tabular*}
\label{tab: fine-tune hyper-parameters}
\end{table*}

\begin{table*}[htb]
\centering
\caption{The table illustrates the pre-training dataset size and model size for BatmanNet and a series of advanced baselines, along with their average AUC across all classification datasets for molecular property prediction.}
\begin{tabular*}{\textwidth}{@{\extracolsep\fill}lccc}
\toprule
Model & Pre-training Data Size (M) & Model Size (M) & AVG-AUC /\%  \\ \midrule
GraphMAE & 2 & - & 78.90\\ 
GROVERbase & 11 & 40 & 82.28 \\ 
GROVERlarge & 11 & 100 & 83.40 \\ 
KPGT & 2 & - & 82.53\\ 
MPG & 11 & 55 & 84.18 \\ 
GEM & 20 & - & 85.15 \\ \midrule
BatmanNet & 0.25 & 2.6& 84.78 \\ 
\bottomrule
\end{tabular*}
\label{tab: model-size}
\end{table*}

\begin{sidewaystable*}[htbp]
\caption{Overall performance for classification tasks and regression tasks of molecular properties prediction following the same experimental settings used in GEM}\label{molecular property prediction-GEM}
\renewcommand\arraystretch{1.2}
\begin{tabular*}{\textheight}{@{\extracolsep\fill}lccccccccc}
\toprule%
\multicolumn{1}{@{}l@{}}{Methods} & \multicolumn{9}{@{}c@{}}{Classification (AUC-ROC)}  \\
\midrule
Dataset & BACE & BBBP & Clin Tox & SIDER & Tox21 & ToxCast & HIV & MUV & Avg \\
\#molecules & 1513 & 2039 & 1478 & 1427 & 7831 & 8575 & 41127 & 93087 & - \\
\#tasks & 1 & 1 & 2 & 27 & 12 & 617 & 1 & 17 & -\\
\midrule
D-MPNN & 0.809$_{(0.006)}$ & 0.710$_{(0.003)}$ & 0.906$_{(0.006)}$ & 0.570$_{(0.007)}$ & 0.759$_{(0.007)}$ & 0.655$_{(0.003)}$ & 0.771$_{(0.005)}$ & 0.786$_{(0.014)}$ & 0.746 \\
AttentiveFP & 0.784$_{(0.022)}$ & 0.643$_{(0.018)}$ & 0.847$_{(0.003)}$ & 0.606$_{(0.032)}$ & 0.761$_{(0.005)}$ & 0.637$_{(0.002)}$ & 0.757$_{(0.014)}$ & 0.766$_{(0.015)}$ & 0.735 \\
N-GramRF & 0.779$_{(0.015)}$ & 0.697$_{(0.006)}$ & 0.775$_{(0.040)}$ & 0.668$_{(0.007)}$ & 0.743$_{(0.004)}$ & - & 0.772$_{(0.001)}$ & 0.769$_{(0.007)}$  & - \\
N-GramXGB & 0.791$_{(0.013)}$ & 0.691$_{(0.008)}$ & 0.875$_{(0.027)}$ & 0.655$_{(0.007)}$ & 0.758$_{(0.009)}$ & - & 0.787$_{(0.004)}$ & 0.748$_{(0.002)}$  & - \\
PretrainGNN & 0.845$_{(0.007)}$ & 0.687$_{(0.013)}$ & 0.726$_{(0.015)}$ & 0.627$_{(0.008)}$ & 0.781$_{(0.006)}$ & 0.657$_{(0.006)}$ & 0.799$_{(0.007)}$ & 0.813$_{(0.021)}$  & 0.742 \\
GROVERbase & 0.826$_{(0.007)}$ & 0.700$_{(0.001)}$ & 0.812$_{(0.030)}$ & 0.648$_{(0.006)}$ & 0.743$_{(0.001)}$ & 0.654$_{(0.004)}$ & 0.625$_{(0.009)}$ & 0.673$_{(0.018)}$ & 0.710\\
GROVERlarge & 0.810$_{(0.014)}$ & 0.695$_{(0.001)}$ & 0.762$_{(0.037)}$ & 0.654$_{(0.001)}$ & 0.735$_{(0.001)}$ & 0.653$_{(0.005)}$ & 0.682$_{(0.011)}$ & 0.673$_{(0.018)}$ & 0.708\\
GraphMAE & 0.831$_{(0.009)}$ & 0.720$_{(0.006)}$ & 0.823$_{(0.012)}$ & 0.603$_{(0.011)}$ & 0.755$_{(0.006)}$ & 0.641$_{(0.003)}$ & 0.772$_{(0.010)}$ & 0.763$_{(0.024)}$ & 0.739\\
GEM & 0.856$_{(0.011)}$ & 0.724$_{(0.004)}$ & 0.901$_{(0.013)}$ & 0.672$_{(0.004)}$ & 0.781$_{(0.001)}$ & 0.692$_{(0.004)}$ & 0.806$_{(0.009)}$ & 0.817$_{(0.005)}$ & 0.781\\
\midrule
BatmanNet & \textbf{0.861}$_{\textbf{(0.028)}}$ & \textbf{0.838}$_{\textbf{(0.005)}}$ & 0.897$_{(0.012)}$ & 0.659$_{(0.003)}$ & \textbf{0.792}$_{\textbf{(0.003)}}$ & \textbf{0.718}$_{\textbf{(0.007)}}$ & \textbf{0.812}$_{\textbf{(0.009)}}$ & 0.784$_{(0.014)}$& \textbf{0.795}\\
\botrule
\end{tabular*}

\begin{tabular*}{\textheight}{@{\extracolsep\fill}lcccc}
\toprule
\multicolumn{1}{@{}l@{}}{Methods} & \multicolumn{4}{@{}c@{}}{Regression (RMSE)} \\
\midrule
Model & ESOL  & FreeSolv & Lipo & Avg \\
\#molecules & 1128 & 642 & 4200  & - \\
\#tasks & 1 & 1 & 1 & - \\
\midrule
D-MPNN & 1.050$_{(0.008)}$ & 2.082$_{(0.082)}$ & 0.683$_{(0.016)}$ & 1.272\\
AttentiveFP & 0.877$_{(0.029)}$ & 2.073$_{(0.183)}$ & 0.721$_{(0.001)}$ & 1.224\\
N-Gram\_RF & 1.074$_{(0.107)}$ & 2.688$_{(0.085)}$ & 0.812$_{(0.028)}$ & 1.525\\
N-Gram\_XGB & 1.083$_{(0.082)}$ & 5.061$_{(0.744)}$ & 2.072$_{(0.030)}$ & 2.739\\
PretrainGNN & 1.100$_{(0.006)}$ & 2.764$_{(0.002)}$ & 0.739$_{(0.003)}$ & 1.534\\
GROVERbase & 0.983$_{(0.090)}$ & 2.176$_{(0.052)}$ & 0.817$_{(0.008)}$ & 1.325\\
GROVERlarge & 0.895$_{(0.017)}$ & 2.272$_{(0.051)}$ & 0.823$_{(0.010)}$ & 1.330\\
GEM   & 0.798$_{(0.029)}$ & 1.877$_{(0.094)}$ & 0.660$_{(0.008)}$ & 1.112\\
\midrule
BatmanNet & \textbf{0.792}$_{\textbf{(0.013)}}$ & \textbf{1.802}$_{\textbf{(0.006)}}$ & 0.729$_{(0.015)}$ & \textbf{1.108} \\
\bottomrule
\end{tabular*}%
\end{sidewaystable*}

\section{Supplementary Section 3: Supplementary Experimental Results}
\subsection{3.1 Additional experiments}
We additionally conduct experiments on molecular properties prediction following the same experimental settings used in GEM \citep{fang2022geometry}. As shown in Table \ref{molecular property prediction-GEM}, BatmanNet achieves state-of-the-art performance on 7 out of 11 datasets, with an overall relative improvement of $1.1\%$ compared to the previous SOTA results on all the datasets ($1.8\%$ on classification tasks and $0.4\%$ on regression tasks). Note that, the results of GraphMAE are from \citep{hou2022graphmae}, and the results of other baselines are directly copied from \cite{fang2022geometry}.

\setcounter{figure}{0}
\renewcommand{\thefigure}{S\arabic{figure}}
\begin{figure*}[htb]
\centering
\includegraphics[width=1.0\textwidth]{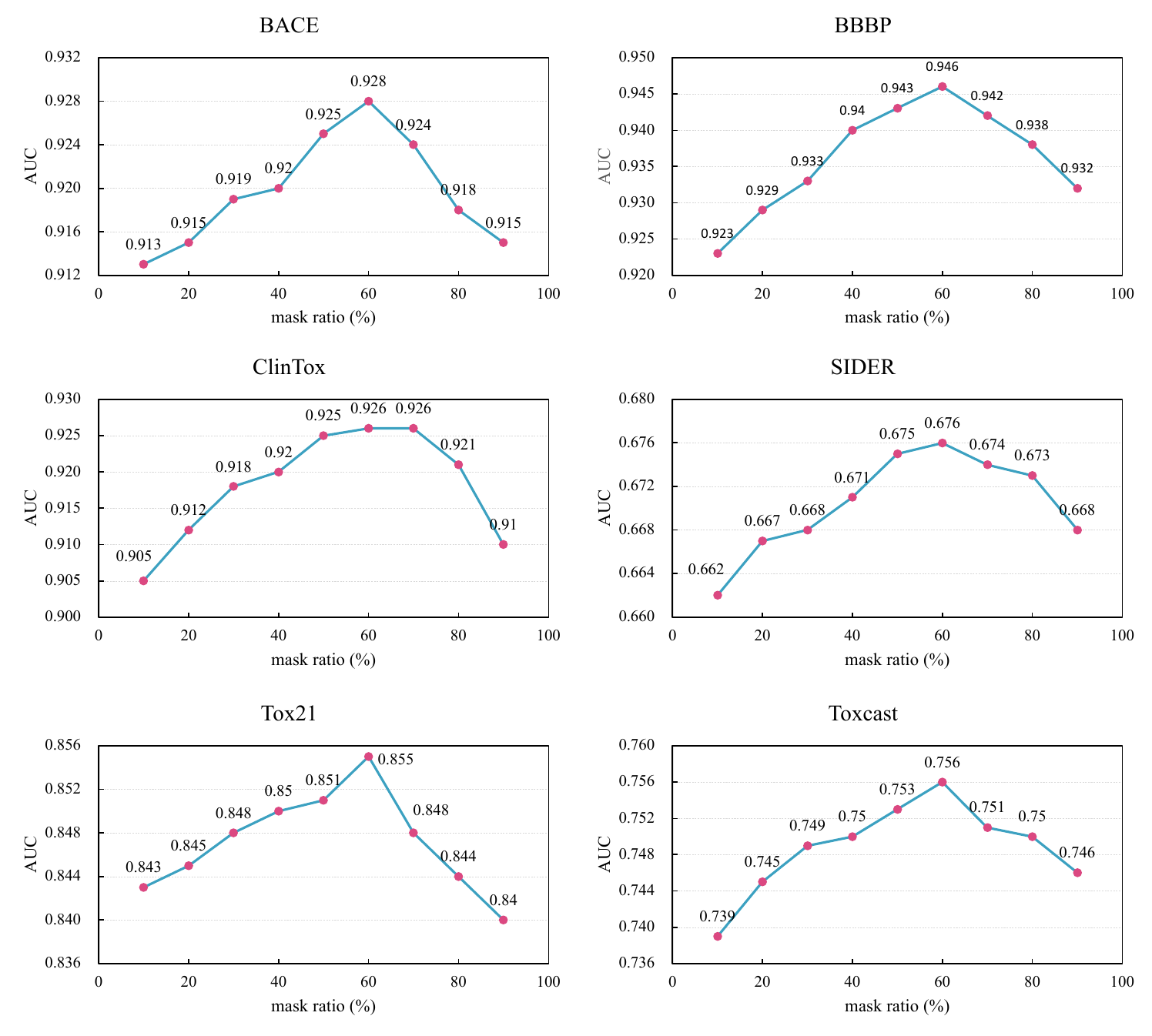} 
\caption{The influence of the masking ratio on each benchmark dataset.}
\label{fig4}
\end{figure*}

\subsection{3.2 Details of The Effect of Different Masking Ratio}
Table \ref{tab: mask ratio} shows the specific experimental results of the BatmanNet pre-trained with different masking ratios (ranging from 0.1 to 0.9) on eight benchmark datasets. Figure \ref{fig4} shows the influence of the masking ratio on each benchmark dataset, respectively. The results show that setting the masking ratio to $60\%$ achieves the best prediction performance, demonstrating the consistency of our BatmanNet's performance on various datasets.

\begin{table*}[htbp] 
\centering
\caption{The experimental results of the BatmanNet pre-trained with different masking ratios (ranging from 0.1 to 0.9) on eight benchmark datasets. We report the mean (and standard deviation) AUC for each dataset of three random seeds with scaffold splitting.}
\begin{tabular*}{\textwidth}{@{\extracolsep\fill}c|cccccc|c}
\toprule
 Ratio & {BBBP} & {SIDER}  & {ClinTox} & {BACE} & {Tox21} & {ToxCast} & {Avg}\\
\midrule
0.1 & $ 0.923_{(0.032)} $ & $\text{0.662}_\text{(0.015)}$ & $\text{0.905}_\text{(0.028)}$ & $\text{0.913}_\text{(0.007)}$ & $\text{0.843}_\text{(0.014)}$ & $\text{0.739}_\text{(0.011)}$ & $\text{0.831}$ \\
0.2 & $\text{0.929}_\text{(0.027)}$ & $\text{0.667}_\text{(0.003)}$ & $\text{0.912}_\text{(0.012)}$ & $\text{0.915}_\text{(0.006)}$ & $\text{0.845}_\text{(0.009)}$ & $\text{0.745}_\text{(0.009)}$ & $\text{0.836}$ \\
0.3 & $\text{0.933}_\text{(0.018)}$ & $\text{0.668}_\text{(0.003)}$ & $\text{0.918}_\text{(0.025)}$ & $\text{0.919}_\text{(0.013)}$ & $\text{0.848}_\text{(0.017)}$ & $\text{0.749}_\text{(0.007)}$ & $\text{0.839}$\\
0.4 & $\text{0.940}_\text{(0.011)}$ & $\text{0.671}_\text{(0.006)}$ & $\text{0.920}_\text{(0.028)}$ & $\text{0.920}_\text{(0.014)}$ & $\text{0.850}_\text{(0.014)}$ & $\text{0.750}_\text{(0.009)}$ & $\text{0.842}$\\
0.5 & $\text{0.943}_\text{(0.019)}$ & $\text{0.675}_\text{(0.004)}$ & $\text{0.925}_\text{(0.025)}$ & $\text{0.925}_\text{(0.014)}$ & $\text{0.851}_\text{(0.013)}$ & $\text{0.753}_\text{(0.008)}$ & $\text{0.845}$\\
0.6 & \textbf{0.946}$_{\textbf{(0.007)}}$ & \textbf{0.676}$_{\textbf{(0.004)}}$ & \textbf{0.926}$_{\textbf{(0.015)}}$ & \textbf{0.928}$_{\textbf{(0.015)}}$ & \textbf{0.855}$_{\textbf{(0.013)}}$ & \textbf{0.756}$_{\textbf{(0.009)}}$ & $\textbf{0.848}$\\
0.7 & $\text{0.942}_\text{(0.008)}$ & $\text{0.674}_\text{(0.004)}$ & \textbf{0.926}$_{\textbf{(0.011)}}$ & 0.924$_{(0.016)}$ & $ 0.848_{(0.012)}$ & $0.751_{(0.007)}$ & $\text{0.844}$\\
0.8 & $\text{0.938}_\text{(0.012)}$ & $\text{0.673}_\text{(0.004)}$ & $\text{0.921}_\text{(0.288)}$ & $\text{0.918}_\text{(0.016)}$ & $\text{0.844}_\text{(0.014)}$ & $\text{0.750}_\text{(0.008)}$ & $\text{0.841}$\\
0.9 & $\text{0.932}_\text{(0.020)}$ & $ 0.668_{(0.005)}$ & $\text{0.910}_\text{(0.022)}$ & $\text{0.915}_\text{(0.015)}$ & $\text{0.840}_\text{(0.015)}$ & $\text{0.746}_\text{(0.011)}$ & $\text{0.835}$\\
\bottomrule
\end{tabular*}
\label{tab: mask ratio}
\end{table*}

\end{appendices}
\end{document}